%% file: neurips_2025.tex
\documentclass{article}

\PassOptionsToPackage{numbers}{natbib}
 \usepackage[preprint]{neurips_2025}

\usepackage[utf8]{inputenc} %
\usepackage[T1]{fontenc}    %
\usepackage{hyperref}       %
\usepackage{url}            %
\usepackage{booktabs}       %
\usepackage{amsfonts}       %
\usepackage{nicefrac}       %
\usepackage{microtype}      %
\usepackage[dvipsnames]{xcolor}         %
\usepackage{graphicx}
\usepackage{xspace}
\usepackage{amsmath}
\usepackage{amssymb}
\usepackage{pifont}
\usepackage{multirow}
\usepackage{subcaption}
\usepackage{changepage}

\title{
VoCap: Video Object Captioning and Segmentation from Any Prompt
}

\author{Jasper Uijlings~\thanks{Equal contribution. Correspondence: jrru@google.com}
 \And Xingyi Zhou$^*$\thanks{Work done while at Google DeepMind} \And Xiuye Gu$^*\dagger$ \And Arsha Nagrani \And Anurag Arnab \And
Alireza Fathi \And David Ross \\ Google DeepMind \And Cordelia Schmid
}

\begin{document}
\input{macros}

\maketitle

\input{0_abstract}    
\input{1_intro}

\input{2_related_work}

\input{3_datasets}

\input{4_model}

\input{5_results}

\input{6_conclusion}
{
    \small
    \bibliographystyle{ieeenat_fullname}
    \bibliography{main}
}

\appendix

\newpage
\include{appendix_arxiv}

\end{document}

%% file: macros.tex
\providetoggle{showcomments}
\settoggle{showcomments}{true} %

\iftoggle{showcomments}{%
    \newcommand{\resolved}[3][]{\ifstrequal{#1}{resolved}{\textcolor{blue}{RESOLVED:}~\textbf{{\MakeUppercase #2:}}~{#3}}{\textbf{\MakeUppercase #2:}~#3}}
    \newcommand{\jasper}[2][]{\textcolor{violet}{\resolved[#1]{jasper}{#2}}}
    \newcommand{\xingyi}[2][]{\textcolor{RoyalPurple}{\resolved[#1]{Xingyi}{#2}}}
    \newcommand{\xingyitext}[1]{\textcolor{RoyalPurple}{{#1}}}
    \newcommand{\jaspertext}[1]{\textcolor{violet}{{#1}}}
    \newcommand{\xiuye}[2][]{\textcolor{Orange}{\resolved[#1]{Xiuye}{#2}}}
    \newcommand{\arsha}[2][]{\textcolor{blue}{\resolved[#1]{Arsha}{#2}}}
    \newcommand{\alireza}[2][]{\textcolor{RoyalPurple}{\resolved[#1]{Alireza}{#2}}}
    \newcommand{\cordelia}[2][]{\textcolor{OliveGreen}{\resolved[#1]{Cordelia}{#2}}}    
    \newcommand{\david}[2][]{\textcolor{RedOrange}{\resolved[#1]{David}{#2}}}
    \newcommand{\changed}[1]{\textcolor{blue}{#1}}
    \newcommand{\anurag}[1]{\textcolor{RedOrange}{[Anurag]: #1}}
}{%
    \newcommand{\changed}[1]{#1}
    \newcommand{\jasper}[2][]{}
    \newcommand{\xingyi}[2][]{}
    \newcommand{\xiuye}[2][]{}
    \newcommand{\arsha}[2][]{}    
    \newcommand{\anurag}[2][]{}    
    \newcommand{\alireza}[2][]{}    
    \newcommand{\cordelia}[2][]{}    
    \newcommand{\david}[2][]{}    
}

\newcommand{\checknum}[1]{\textcolor{RedOrange}{#1}}
\newcommand{\para}[1]{\noindent\textbf{#1}}
\newcommand{\vocap}{VoCap\xspace}
\newcommand{\savs}{SAV-Caption\xspace}
\newcommand{\savseval}{SAV-Caption-val\xspace}
\newcommand{\savstrain}{SAV-Caption-train\xspace}
\newcommand{\na}{\ding{55}}

\newcommand{\refsec}[1]{Sec.~\ref{sec:#1}}
\newcommand{\refapp}[1]{Appendix~\ref{sec:#1}}
\newcommand{\refthm}[1]{Theorem~\ref{thm:#1}}
\newcommand{\refcly}[1]{Corollary~\ref{cly:#1}}
\newcommand{\reftbl}[1]{Tab.~\ref{tab:#1}}
\newcommand{\reffig}[1]{Fig.~\ref{fig:#1}}
\newcommand{\refalg}[1]{Alg.~\ref{alg:#1}}
\newcommand{\refline}[1]{Line~\ref{line:#1}}
\newcommand{\shortrefsec}[1]{\S\ref{sec:#1}}
\newcommand{\refeq}[1]{Eq.~\ref{eq:#1}}
\newcommand{\refeqshort}[1]{\eqref{eq:#1}}
\newcommand{\shortrefeq}[1]{Eq.~\eqref{eq:#1}}
\newcommand{\lblfig}[1]{\label{fig:#1}}
\newcommand{\lblsec}[1]{\label{sec:#1}}
\newcommand{\lbleq}[1]{\label{eq:#1}}
\newcommand{\lblcly}[1]{\label{cly:#1}}
\newcommand{\lblthm}[1]{\label{thm:#1}}
\newcommand{\lbltbl}[1]{\label{tab:#1}}
\newcommand{\lblalg}[1]{\label{alg:#1}}
\newcommand{\lblline}[1]{\label{line:#1}}

%% file: 0_abstract.tex
\begin{abstract}
Understanding objects in videos in terms of fine-grained localization masks and detailed semantic properties is a fundamental task in video understanding. In this paper, we propose VoCap, a flexible video model that consumes a video and a prompt of various modalities (text, box or mask), and produces a spatio-temporal masklet with a corresponding object-centric caption. As such our model addresses simultaneously the tasks of promptable video object segmentation, referring expression segmentation, and object captioning. Since obtaining data for this task is tedious and expensive, we propose to annotate an existing large-scale segmentation dataset (SAV) with pseudo object captions. We do so by preprocessing videos with their ground-truth masks to highlight the object of interest and feed this to a large Vision Language Model (VLM). For an unbiased evaluation, we collect manual annotations on the validation set. We call the resulting dataset SAV-Caption. We train our VoCap model at scale on a SAV-Caption together with a mix of other image and video datasets. Our model yields state-of-the-art results on referring expression video object segmentation, is competitive on semi-supervised video object segmentation, and establishes a benchmark for video object captioning. 
Our dataset is available at \url{https://github.com/google-deepmind/vocap}.

\end{abstract}

%% file: 1_intro.tex
\section{Introduction}\label{sec:introduction}

Understanding objects in videos, including both their fine-grained locations (represented as segmentation masks) as well as their detailed semantic properties, is a fundamental task in video understanding.
It serves as a basic block for various applications, including video generation and editing~\cite{chai2023stablevideo, hu2024instruct,wang2024instancediffusion}, wild animal care~\cite{beery2020context,sun2024video}, and self-driving~\cite{caesar2020nuscenes,sun2020scalability}.
While it is trivial for a human to point to an object in a video and describe it in detail, there is yet no existing computer vision system that is capable of both \textit{spatio-temporal} localization {via segmentation masks}, as well as a \textit{semantic} understanding of objects via natural language.

\input{figures/framework}

In this paper, we propose a model and data for fine-grained video object understanding with flexible inputs and outputs modalities.
Our model consumes a video and an input prompt, where the prompt can be a mask and box, but also natural language (i.e. referring expression). Our model then produces \textit{both} a spatio-temporal mask (i.e., a ‘masklet’) \textit{and} a free-form natural language caption describing the object.  %
Because the output caption is a free-form sentence, it can describe the attributes of the object as well as how they change over time.
Our model can be used for a variety of tasks bridging localization and language, for example referring object segmentation~\cite{yu2016modeling,seo2020urvos} or location-conditioned captioning~\cite{krishna2017visual}, which we extend to video.

Several previous works attempt to bridge this gap between visual localization and language understanding - for example segmentation via free-form referring expressions~\cite{khoreva2018rvos,seo2020urvos,wu2022language}, where the goal is to produce a segmentation mask for an object given a short description which refers to a single object, or dense video object captioning (DenseVOC) ~\cite{zhou2023dense}, which produces bounding boxes and captions for all classes within a certain vocabulary in a video.
While localization with referring expressions~\cite{khoreva2018rvos,seo2020urvos,wu2022language} which typically only takes in a minimal-required text to identify an object in the input, our model can also produce detailed captions given a location prompt.
Unlike DenseVOC~\cite{zhou2023dense} -- which is non-promptable, is trained on a fixed set of objects, and which is limited to producing boxes only -- our model works with flexible input prompts and produces dense masklets as output in addition to captions.
Our model is inspired by both existing captioning Vision-language models (VLMs) such as BLIP2~\cite{li2023blip}, as well as promptable segmentation models such as SAM2~\cite{ravi2024sam}, and brings a number of related video segmentation and captioning tasks together while enabling cross-task synergies.
Specifically, on top of the general SAM2 design~\cite{ravi2024sam} we introduce a lightweight shared text encoder and decoder based on BERT~\cite{devlin2018bert}, and an efficient caption feature extractor similar to QFormer~\cite{li2023blip}.
These modules are pre-trained on large-scale vision-language datasets~\cite{chen2022pali}, and are used to encode the text prompt and decode the output caption.
This results in a unified promptable model for Video Object Captioning and Segmentation from Any Prompt (VoCap) that takes as input a prompt (text, mask, or box) and outputs segmentation mask and caption jointly (See Fig.~\ref{fig:framework}).

Obtaining data to train our VoCap model is a significant challenge -- annotating video with segmentation masklets and captions is tedious and expensive, and not easily scalable to large volumes of data.
Hence we propose a pseudo-labeling pipeline starting with the SAV Manual dataset~\cite{ravi2024sam}, which contains accurate segmentation masks. %
We then automatically generate \textit{object-centric} captions using a large-scale VLM (Gemini 1.5 Pro Vision~\cite{gemini2024pro1.5}).
By pre-processing the videos to highlight each object mask and blur the background, we steer the VLM to describe each object and what happens to it with satisfying accuracy and details. This enables us to generate a large-scale training set with masks and object-centric captions without additional human labor.
We then combine this dataset with existing partially annotated datasets~\cite{krishna2017visual,seo2020urvos,yu2016modeling,chen2022pali} to co-train our model. 

For evaluation, we ran a \emph{human} annotation campaign on the SAV-val dataset~\cite{ravi2024sam} where each object is captioned by three different annotators.
Furthermore,
we evaluate our model on existing datasets and tasks such as mask-prompted video segmentation on MOSE~\cite{ding2023mose} and referring segmentation on RefVOS~\cite{seo2020urvos}.
To summarize, we make the following contributions:
\begin{itemize}
\item We present a unified promptable model VoCap that can produce both spatio-temporal masklets and captions for objects in video. Our model is flexible in both the input and the output, taking as input a prompt (text, mask, or box) and outputting masklets and captions.

\item We collect manually annotated object captions on SAV-val and create pseudo-captions by leveraging existing mask annotations on SAV-train using Gemini Pro 1.5. We make both the manual and pseudo-annotations publicly available~\footnote{\url{ 
https://github.com/google-deepmind/vocap/
}}.  %
By showing good performance on the manually annotated object captions, we demonstrate that these pseudo-labels are effective for training our captioning model.

\item We set a new state-of-the art for Referring Expression Video Object Segmentation, show competitive results on semi-supervised object segmentation, and establish a benchmark for video object captioning.

\end{itemize}

%% file: figures/framework.tex
\begin{figure*}[!t]
\vspace{-.7cm}
	\center
 \includegraphics[width=0.95\linewidth]{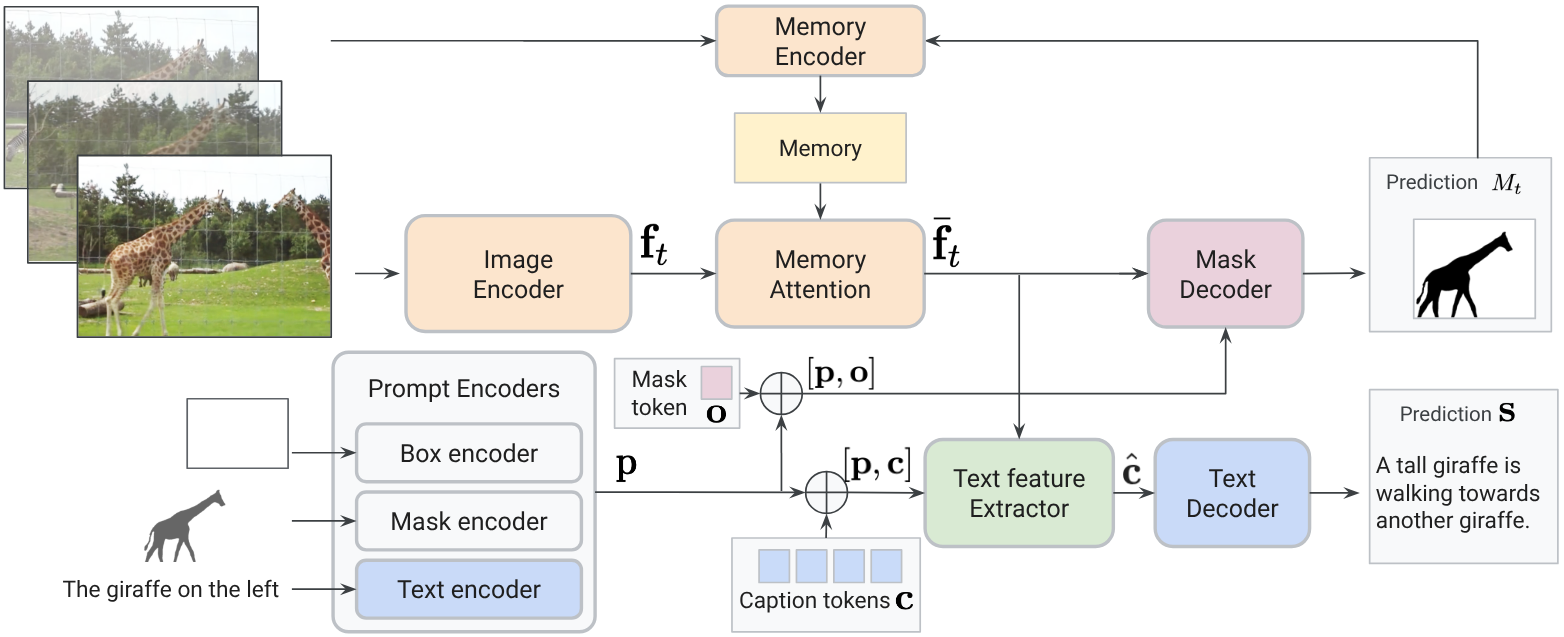}
	\caption{
 \small{\textbf{Overview of our \vocap architecture.} 
 Our model processes videos frame-by-frame, with access to an updating memory for each object.
 Each frame goes through the image encoder, cross attends to the memory.
 The memory-aggregated image features and the object-specific prompt embeddings are fed into the mask decoder to obtain the mask predictions.
 The memory module is updated with the per-frame mask predictions and image features.
 In addition, our model also takes text prompts, and we use a novel text feature extractor and text decoder to produce captions for the object. The text encoder and text decoder share the architecture and weights.
 }}
 \label{fig:framework}
 \vspace{-.6cm}
\end{figure*}

%% file: 2_related_work.tex
\section{Related Work}\label{sec:related_work}

\para{Segmentation and Captioning Models.}
A variety of models are dedicated to video segmentation and expect an initial input 
mask~\cite{yang2021aot,yang2021aost,yang2022deaot,cheng2022xmem,cheng2024putting,guo2024videosam,deng2024memsam,ravi2024sam,yang2023track}, a referring expression~\cite{seo2020urvos,lan2023referring,wu2022language}, or both~\cite{cheng2023segment,wu2023uniref++}. CLIPSeg~\cite{luddecke2022cvpr_clipseg} can also consume a query image.
SAM2~\cite{ravi2024sam} can do segmentation without any inputs as is done in DEVA~\cite{cheng2023segment} by starting from the original SAM~\cite{kirillov2023segment} with a point grid prompts.
However, none of these models can generate descriptions.
In captioning, there are works on \emph{global video} captioning which describe the whole video~\cite{kanani2021ijcnn_videodescriptions,iashin2020bmvc_vidcapbimodal,yang2023cvpr_vid2seq,yao2015iccv_videocaption,wang2021iccv_densevidcap}, or on \emph{dense image} captioning which provide object-centric captions and their locations in images~\cite{johnson2016cvpr_densecap,li2019aaai_densecap,shao2022nnls_densecap,zhang2023gpt4roi,yuan2024osprey,peng2023kosmos,xu2024cvpr_pixelllm}.
Only few works do \emph{dense video} captioning~\cite{choudhuri2024ow,zhou2023dense} for objects.
DenseVOC~\cite{zhou2023dense} predicts bounding boxes with captions but does not predict masks and only detects a predefined set of object classes.
The OW-VISCapTor model~\cite{choudhuri2024ow} predicts segments with captions, but cannot handle textual or mask input prompts.
Furthermore, OW-VISCapTor is based on an image-first \emph{tracking-by-detection} paradigm that can be suboptimal in long videos with occlusions,
while we build on top of strong \emph{memory-based} trackers~\cite{ravi2024sam,cheng2022xmem} and can handle long and challenging videos~\cite{ding2023mose}

\noindent \para{Datasets.} While numerous datasets exist for video object segmentation and captioning separately, very few combine both on the same set.
Video segmentation datasets include various input forms, including a mask given on the first frame (\textit{semi-supervised video object segmentation}, or \textit{SS-VOS})~\cite{perazzi2016davis,caelles2019davis,ding2023mose,qi2022ijcv_ovis,wang2021uvo}, a target object class (\textit{semantic object segmentation})~\cite{kim2020cvpr_panopticvideo,real2017youtubeboundingboxes,russakovsky2015imagenet} or a referring expression~\cite{ding2023mevis,khoreva2018rvos,seo2020urvos,wu2022language} (\emph{Referring Video Object Segmentation}, or \emph{RefVOS}).
While referring expression datasets have masks and referring text, this text tends to be mainly focused on identifying the object in the video, not describing it.
To link captions better to the visual domain, several datasets focus on having \emph{grounded} captions in both the image domain~\cite{krishna2017visual,lin2024draw,plummer2015cvpr_flickr30k,peng2023kosmos,ponttuset2020eccv_locnar,wang2023all,wang2024all,xue2024xgen} and video domain~\cite{voigtlaender2023connecting,zhang2020does,zhou2018youcook2bb,zhou2019grounded}.
In particular, in the video domain, 
bounding box annotations are added 
in~\cite{zhou2018youcook2bb} to YouCook2~\cite{zhou2018youcook2},
and in~\cite{zhou2019grounded} to ActivityNet~\cite{krishna2017dense}. 
In~\cite{zhang2020does} the relations of VidOR~\cite{shang2019icmr_vidor} are converted into captions while grounding is provided by the existing bounding boxes. BenSMOT~\cite{li2024eccv_bensmot} provides a human-focused dataset with boxes, their object-centric captions, and interactions.
Video Localized Narratives~\cite{voigtlaender2023connecting} introduced an object-centric protocol in which captions are grounded by a mouse trace.
In contrast, in this paper we provide a stronger form of grounding by linking captions to \emph{segmentation masks}. 
Our pseudo-labeled dataset is also an order of magnitude larger than these datasets (see Table \ref{tab:dataset_comparison}).

\noindent \para{Pseudo labels.} With increasing model capabilities and increasing data requirements for training large models, it is increasingly common to use automatically generated labels in the pre-training stage. 
SAM2~\cite{ravi2024sam} provides automatically generated masks on their SAV dataset, enabling distillation. 
BLIP3~\cite{xue2024xgen}, OWLv2~\cite{minderer2023neurips_owlvit2}, and Kosmos-2~\cite{peng2023kosmos} go beyond distillation for bounding box generation by exploiting existing captions: they extract noun phrases from the captions, feed them to an open-world detector, and only keep high-scored boxes.
MVDP~\cite{lin2024draw} draws existing object classes and location annotations in an image using set-of-masks~\cite{yang2023arxiv_setofmark} and feeds this to GPT-4V~\cite{gpt4vision} to generate object-centric captions, relationships, and Q\&A pairs.
In this paper, we augment videos with ground-truth segmentation masks and prompt vision-language models to create high-quality pseudo captions.

%% file: 3_datasets.tex
\section{The SAV-Caption Dataset}\label{sec:datasets}

We want to have a large-scale training set with spatio-temporal segmentation masks and their captions.
Therefore we start from SAV~\cite{ravi2024sam}, 
the largest and most diverse video dataset with segmentation masks.
We use the `Manual' part which was annotated by combining 
SAM2 predictions~\cite{ravi2024sam} with human annotator corrections to ensure high-quality masklets. 
Next we detail how we add captions to the existing SAV segmentation dataset using automatic annotations (Sec.~\ref{sec:automatic_annotation}) and human annotations (Sec.~\ref{sec:human_annotation}). In both cases we want to have captions with the object class, its visual properties (which aligns the captions with visual referring expressions), and what it does (which captures the temporal semantics).
Such captions are aligned with previous dense video captioning datasets (e.g.~\cite{voigtlaender2023connecting,zhang2020does}).

\input{tables/dataset_comparisons}

\subsection{Automatically Annotated Training Data}\label{sec:automatic_annotation}

We use Gemini 1.5 Pro Vision~\cite{gemini2024pro1.5} to automatically generate captions on this dataset.
This model is a long-context vision language model and is therefore suited to consume relatively large video clips ($\approx 1000$ frames).
To create accurate captions, we draw inspiration from works which augment images with visual prompts to focus the attention of the visual models to what matters, thereby simplifying the task~\cite{nasiriany2024icml_pivot,yang2023arxiv_setofmark,zheng2024icml_grounded_gpt,wu2024eccv_dettoolchain, shtedritski2023does}.
In particular, we adopt two visual prompting techniques: 1) We highlight the target segment by drawing a clear red contour around it (\textbf{Contour});
2) upon finding that Gemini would still sometimes focus on objects in the background, we blurred the background using a Gaussian filter (\textbf{Blur}).
Both modifications are explicitly mentioned in the textual prompt.
An example of the video frame we fed to Gemini can be seen in ~\reffig{prompt}.

For the textual part of the prompt, we carefully iterated to increase the quality of the generated caption.
In this process, we found it helpful to structure the prompt: we ask to describe first the object, then its visual properties, and then what it does, and finally we ask it to give the caption while keeping earlier mentioned elements consistent.
Statistics of SAV-Caption train are given in Tab.~\ref{tab:dataset_comparison},
and example captions are shown in~\reffig{gemini_train_visu}.
The exact prompt and a quantitative analysis is given in the supplementary.

\subsection{Human Annotated Validation Data}\label{sec:human_annotation}

Our evaluation should be free of any potential biases of any Visual Language Model. Therefore we collect our evaluation set fully manually with three captions per object. 
In particular, we start from SAV-val and instruct the raters to provide a single free-form caption of the object highlighted in the video.
Like in Sec.~\ref{sec:automatic_annotation} we highlight the object with a red border but we do not blur the background.
We have explicit instructions for the annotators to include in their caption the object class, its visual properties, and what it does.
We also ask them to not mention irrelevant objects in the background.
The statistics of SAV-Caption val are given in Tab.~\ref{tab:dataset_comparison}. The annotation instructions and the UI can be found in the supplementary material.

\subsection{Comparison with Other Datasets}\label{sec:dataset_comparison}

There only exist few video datasets where objects are annotated with both spatio-temporal masklets and captions ~\cite{ding2023mevis,khoreva2018rvos,seo2020urvos}. These existing datasets were all made for referring expression segmentation but can be repurposed for the captioning task.
However, referring expressions were made with the intention for objects to be uniquely identifiable, not for semantic understanding.
Furthermore, our training set is at least one order of magnitude bigger.

\input{figures/prompt_and_caption_illustrations}

%% file: tables/dataset_comparisons.tex
\begin{table}[t!]
    \caption{\textbf{Video datasets with masks and captions.} Our SAV-Caption is an order of magnitude larger.}
    \label{tab:dataset_comparison}
    \centering
    \begin{tabular}{@{}lcccccc@{}}
    \toprule
        dataset & \# videos & \# objects captioned & \# words per caption \\ 
        \midrule
        RefVOS-DAVIS~\cite{khoreva2018rvos}     & 150 & 436 & 6.4 \\
        MeVIS~\cite{ding2023mevis}          & 2.0k & 8.1k & 7.3 \\
        RefVOS-YTVOS~\cite{seo2020urvos} & 4.0k & 7.5k & 9.7 \\ 
        \midrule
        SAV-Caption val (manual)            & 155 & 290 & 13.5 \\
        SAV-Caption train (automatic)       & 50k & 170k & 11.8 \\ 
    \bottomrule
    \end{tabular}
\end{table}

%% file: figures/prompt_and_caption_illustrations.tex
\begin{figure}[!t]
\vspace{-.7cm}
    \centering
    \begin{minipage}{0.38\textwidth} %
        \centering
        \includegraphics[width=\linewidth]{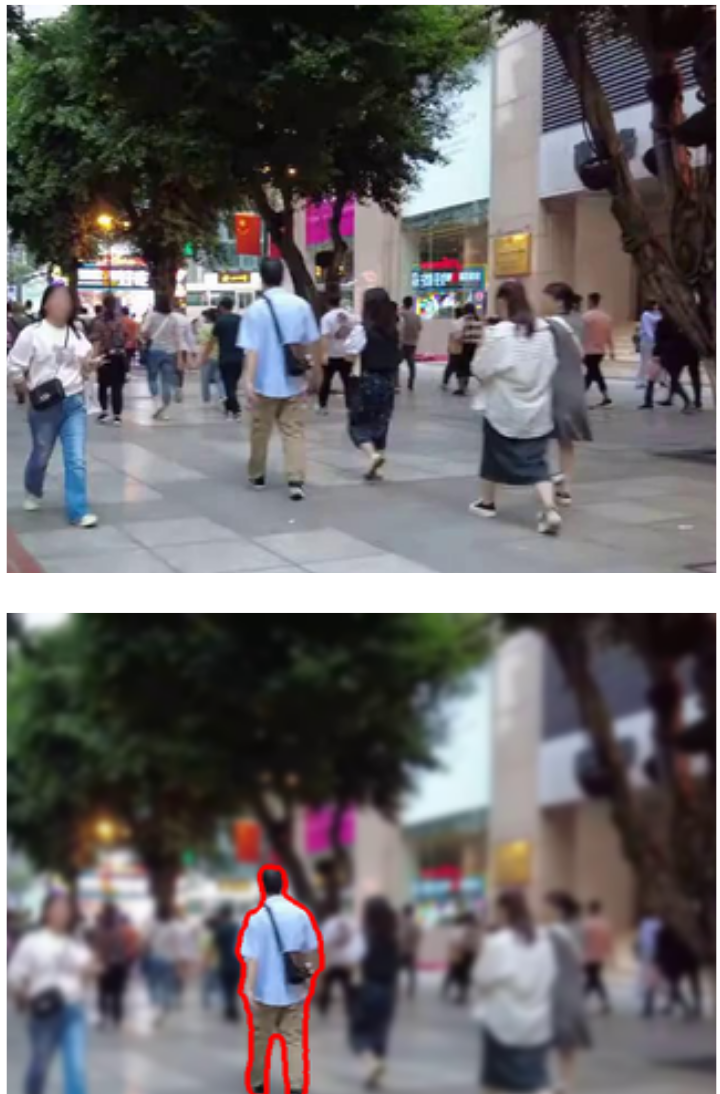}
        \caption{\small{\textbf{Illustration of our visual prompting.} Top: the original frame; Bottom: our processed input to the Gemini annotator. We apply a red contour to highlight the target object and blur the background avoid distractions.}}
        \label{fig:prompt} %
    \end{minipage}
    \hfill %
    \begin{minipage}{0.58\textwidth} %
        \centering
        \includegraphics[width=\linewidth]{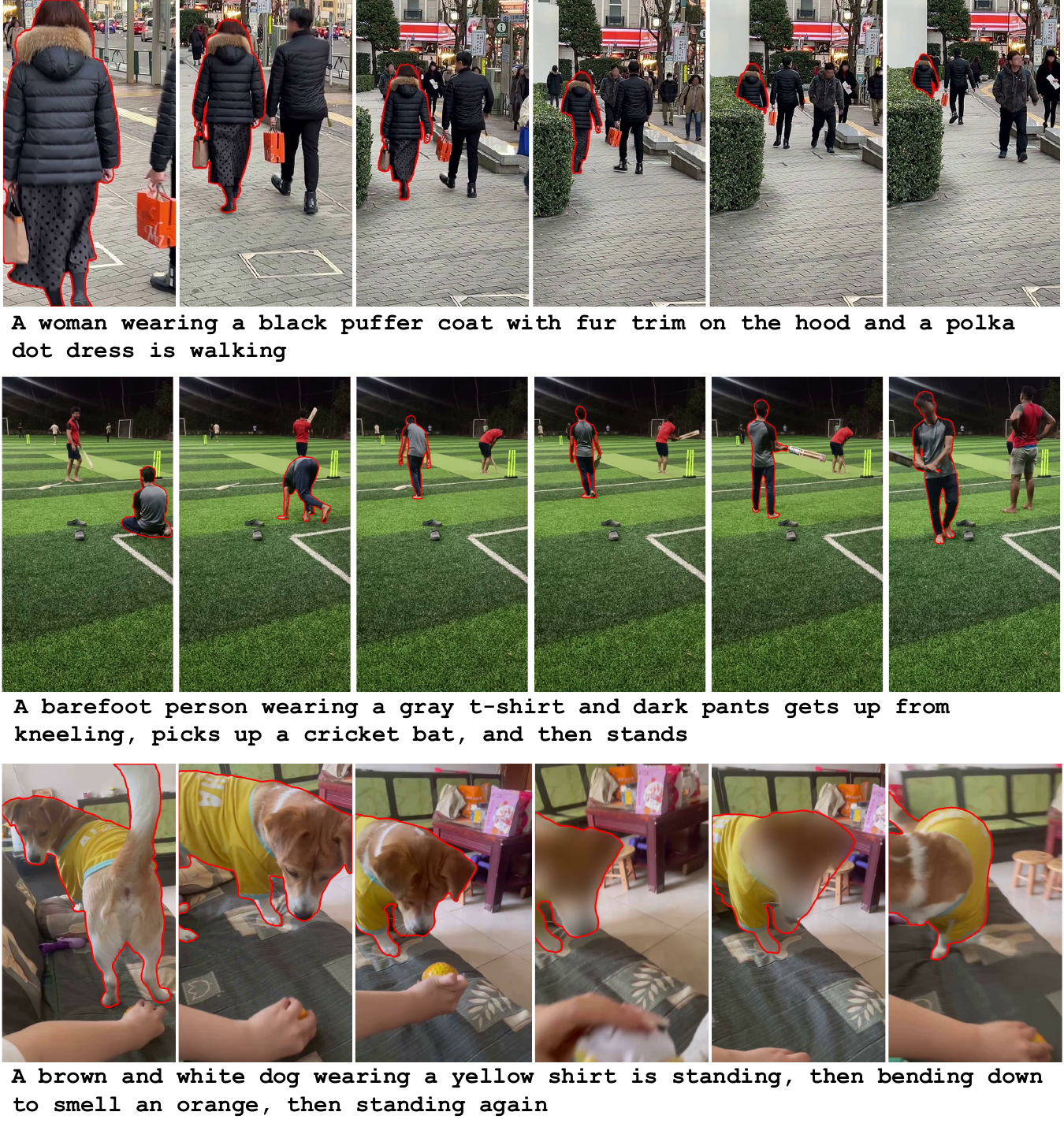}
        \caption{\small{\textbf{Examples of our VLM pseudo-labeled SAV-Caption training set.} We show the target object in red contour, and show the output captions below. The captions capture object classes, appearance properties, and multiple actions connected by ``then''.}}
        \label{fig:gemini_train_visu} %
    \end{minipage}
    \vspace{-3mm} %
    \label{fig:combined_figures} %
\end{figure}

%% file: 4_model.tex
\section{VoCap Model}\label{sec:model}

Given an image or video $V \in \mathbb{R}^{T \times H \times W \times 3}$ (for images, $T=1$) and a prompt, where the prompt can be a bounding box or a mask in the first frame, or a textual description,
our \vocap{} model produces a binary masklet $M \in \mathbb{R}^{T \times H \times W}$ and a caption string $\mathbf{s}$ for the corresponding object.

\subsection{Model Architecture}
As illustrated in~\reffig{framework},
our model is composed of segmentation modules inspired by~\cite{ravi2024sam}, including an image encoder, a memory encoder, a memory attention module, a location prompt encoder, and a mask decoder.
We add new language modules: a text encoder, a text feature extractor, and a text decoder.
As a result, our model can take both texts or masks as inputs or as outputs.

The \textbf{image encoder}
takes a single frame $V_t$ as input and produces down-sampled image features $\mathbf{f}_t \in \mathbb{R}^{H' \times W' \times d}$.
This can be any visual backbone, and we use eva02~\cite{EVA02} given its dedicated pretraining for both language~\cite{radford2021learning} and localization tasks~\cite{he2022masked}.
Following ViTDet~\cite{li2022exploring}, we use simple convolutional upsampling layers~\cite{ronneberger2015u,zheng2021rethinking} to produce multi-scale features as additional inputs for the mask decoder~\cite{ravi2024sam}.
Note that each frame is processed separately without temporal communication.

The \textbf{memory encoder} and \textbf{memory attention} together augment the per-frame image feature $\mathbf{f}_t$ with temporal information.
Specifically, at each timestamp, the memory encoder fuses the input image and output mask into a memory feature, which is stored in a memory bank that keeps a history of {$d'$-dimensional} spatio-temporal appearance features (memory dimension $d'$ can be different from feature dimension $d$).
Following SAM2~\cite{ravi2024sam} we use a fixed-sized memory bank with a first-in-first-out memory queue. 
There are several cross attention layers between the current image feature and the memory bank which makes the output image features $\bar{\mathbf{f}}_t  \in \mathbb{R}^{H' \times W' \times d}$ temporally-aware. 

The \textbf{location prompt encoder} projects location inputs to embeddings.
Specifically, box prompts are encoded as sparse embeddings $\mathbf{p} \in \mathbb{R}^{n \times d}$, where $n$ is the number of points ($n=2$ for 2 box-corners) and $d$ is the feature dimension.
Mask prompts are encoded as dense embeddings $\mathbf{m} \in \mathbb{R}^{H' \times W' \times d}$ with the same shape as the image feature.

The \textbf{text encoder} takes text strings as inputs and projects them to embeddings.
It can be any language model~\cite{devlin2018bert,team2024gemma,team2024gemma2,raffel2020exploring,touvron2023llama} that encodes the integer vocabulary indexes to embeddings.
Specifically, we feed text prompts as the text prefix to the language model with full attention, and extract the features before the vocabulary classification layer.
We use an additional dimension-matching layer to project from the language model embedding space to the prompt embedding space.
We reuse our sparse embedding notation  $\mathbf{p} \in \mathbb{R}^{n \times d}$ for text prompts.
Here $n$ is the number of tokenized words in the text query.
Because the text prompt provides conditioning for the entire video and because the target object does not always appear in the early frames of the video, we feed the text prompt embedding to all frames of the video.

The \textbf{mask decoder} takes the temporal-aware image feature $\bar{\mathbf{f}}_t$ and the prompt features $\mathbf{p}$ or $\mathbf{m}$ as inputs, and outputs the mask at the current frame $M_t \in \mathbb{R}^{H \times W}$.
In SAM, the mask decoder uses cross attention to communicate the image and prompt features:
\begin{align}
\lbleq{cross-attention}
{\tilde{\mathbf{f}}_t}, [\tilde{\mathbf{p}}, \tilde{\mathbf{o}}] & = CA_{seg}(\bar{\mathbf{f}}_t + \mathbf{m}, [\mathbf{p}, \mathbf{o}]) \\
m_t & = \mathcal{D}({\tilde{\mathbf{f}}_t}, \tilde{\mathbf{o}})
\end{align}
where $\mathbf{o} \in \mathbb{R}^{1 \times d}$ is a learned mask token and is concatenated with the sparse prompt $\mathbf{p}$, and $CA$ is the cross-attention operation. $\mathcal{D}$ is a mask decoding function with upsampling convolutions and a final dot-product~\cite{cheng2021per}.
The output of the cross-attention, $\tilde{\mathbf{o}} \in \mathbb{R}^{H \times W}$, can be considered as the object feature conditioned on the prompt.
Besides the mask, the mask decoder also predicts for each frame a binary object appearance indicator $a_t \in \{0, 1\}$ to handle occlusion or out-of-view movement, and an IoU prediction $iou_t$ which estimates the quality of the mask.

\noindent \textbf{Text feature extractor}.
Similar to how the object features are extracted in the mask decoder in~\refeq{cross-attention}, we use learned caption tokens $\mathbf{c} \in \mathbb{R}^{l \times d}$ and cross attention to extract caption features for each object:
\begin{equation}
\hat{\mathbf{f}}, [\hat{\mathbf{p}}, \hat{\mathbf{c}}] = CA_{cap}(\bar{\mathbf{f}}_t + \mathbf{m}, [\mathbf{p}, \mathbf{c}])
\end{equation}
where we only use output $\hat{\mathbf{c}}$ and discard $\hat{\mathbf{f}}$ and $\hat{\mathbf{p}}$. This formulation is analogous to popular vision-feature extractors in vision-language models~\cite{jaegle2021perceiver,alayrac2022flamingo,ryoo2021tokenlearner,li2023blip},
while we additionally condition on the prompt embeddings $\mathbf{m}$ or $\mathbf{p}$.
Following BLIP2~\cite{li2023blip}, we use $l=32$ tokens for the caption tokens.

\noindent \textbf{Text decoder}.
Following popular vision-language model design~\cite{li2023blip,wang2022git,liu2023llava} we feed the object-aware caption feature $\hat{\mathbf{c}}$ as prefix to an auto-regressive language model $\mathcal{L}$ to produce object caption $\mathbf{s}$:
\begin{equation}
\mathbf{s}_i = \mathcal{L}(\hat{\mathbf{c}}, \mathbf{s}_{1:i-1})
\end{equation}
Again, the text decoder can be any language model~\cite{devlin2018bert,team2024gemma,team2024gemma2,raffel2020exploring,touvron2023llama} with a causal attention mask.
We note that both the architecture and the weights of the text encoder and text decoder can be shared even though the text decoder uses causal attention, and the text encoder uses bidirectional attention.
Therefore, during training, the language model is updated for both text encoding and decoding
regardless of whether we use text as an input prompt or as a target output caption.
We follow the standard transformer decoder~\cite{vaswani2017attention,devlin2018bert} as it is simple and effective~\cite{wang2022git,wu2024grit,zhou2023dense}.

\subsection{Training}
Given our flexibility on inputs and outputs, our model can leverage a variety types of annotations from different datasets: For SAV-Caption our model consumes a mask prompt and calculates the loss on both the predicted masklet and the predicted caption. On VisualGenome~\cite{krishna2017dense} our model consumes a box prompt and calculates the loss on the caption. For SS-VOS we have a first frame mask input prompt and a loss on the masklet. For RefVOS we have a text input prompt and a loss on the masklet.
Following other joint models for image and video, we treat images as a single-frame video~\cite{ravi2024sam, villegas2022phenaki, bain2021frozen}.
Concretely, we do not use the memory module (specifically, for $t=0$, $\bar{\mathbf{f}}_0 \equiv \mathbf{f}_0$) for the first frame or images.
To leverage all available data, we first pre-train our language and vision components separately, then perform multi-task training with joint mask- and caption-annotations. Finally, for achieving the best performance, we finetune on specific datasets. See details in~\ref{sec:implementation_details}.

\subsection{Inference}
\lblsec{inference}
Our model runs on images or videos of arbitrary lengths.
Like in training, for an image or the first frame of the video, the visual features are not modified by the memory attention since there are no memories
(again, $\bar{\mathbf{f}}_0 \equiv \mathbf{f}_0$).
For the following frames of the video, our model runs in an online manner: in each frame the model produces both mask and caption outputs, and updates the memory.
Using the object appearance prediction results, we keep the captions where the object exists in the frame, and take the most common caption prediction as the final caption for the trajectory.

%% file: 5_results.tex
\section{Experiments}\label{sec:results}

\subsection{Implementation Details}\label{sec:implementation_details}
We implement our model in JAX~\cite{jax2018github}.
Our image encoder is EVA02-L~\cite{EVA02}, a 24-layer vision transformer (ViT)~\cite{dosovitskiy2020image} with MAE~\cite{he2022masked} and CLIP~\cite{radford2021learning} pretraining.
We chose this encoder as it is more suitable for language tasks, compared to the MAE-pretrained ViT used in SAM~\cite{kirillov2023segment, ravi2024sam}.
Our shared language encoder and decoder is a 6-layer BERT model~\cite{devlin2018bert} with random initialization, which has been shown to be effective and efficient for object captioning in several works~\cite{wu2024grit,zhou2023dense,wang2022git}.
The text feature extractor contains 2 cross-attention layers with the same architecture as the mask decoder.
Other modules follow the SAM2~\cite{ravi2024sam} architecture and are all randomly initialized. In appendix~\ref{sec:sam2_replication} we show that our re-implementation of SAM2 is comparable to the original, and that EVA02-L is a strong alternative backbone.

\input{tables/datasets}

We have three training phases: (i) a \emph{pre-training phase} to initialize weights of the visual and language modalities separately. (ii) A \emph{multi-task training phase} in which we train our full model end-to-end jointly on the three tasks, namely Captioning, SS-VOS, and RefVOS. (iii) A finetuning phase in which we optimize the model per dataset.
In more detail, for the pre-training phase (i) we use the text encoder and decoder of an existing checkpoint trained for image captioning on WebLI~\cite{chen2022pali}.
Since we use an eva02 image backbone, we cannot use the existing SAM2 checkpoints. Instead, we pre-train our visual components on %
SAV~\cite{ravi2024sam}, YTVOS~\cite{xu2018youtube}, and DAVIS~\cite{perazzi2016davis}, 
following the SAM2~data mixture ratio (49.5: 9.2: 1.3).
We train 300k iterations, using a batch size 64 at $512\times512$ resolution.
We verified that this training recipe produces results close to the official SAM2 model which is trained on proprietary datasets and uses a larger resolution (see supplementary for more details).

In our multi-task training phase (ii), we use datasets with both language and segmentation annotations (Tab.~\ref{tab:training_datasets}): VisualGenome~\cite{krishna2017visual}, RefCOCO~\cite{yu2016modeling,mao2016generation,mao2016generation}, RefVOS-YTVOS~\cite{seo2020urvos}, and our \savstrain, with a data mixture ratio 0.5: 1: 1: 2.
We train for 240k iterations with batch size 32 using a $512\times512$ input resolution.
The multi-task training phase takes $\approx 50$ hours on 32 H100 GPUs. 
Since most prior work report numbers specialized per dataset, we have a small finetuning stage per dataset.

\subsection{Captioning}\label{sec:results_captioning}

The localized captioning task is defined as producing a text caption given a location prompt (e.g. box or mask). 
For images we perform image captioning on Visual Genome given a box around an object as input prompt.
For videos we address localized object captioning with when prompted by a mask annotation for the first video frame.
For both the image and captioning tasks we use the standard CIDEr~\cite{vedantam2015cider} metric.

\para{Video Object Captioning Baselines.}
Since our VoCap model is the first model which can do simultaneous object segmentation and captioning given a first-frame input mask, there are no existing methods to which compare to. However, we present results for a few strong baselines. First, we run a semi-supervised VOS method to obtain segments, and feed these into existing off-the-shelf captioning models. In particular, we run our re-implementated and retrained SAM2 model~\cite{ravi2024sam} as the SS-VOS method and apply the popular captioning models BLIP2~\cite{li2023blip} (which predicts captions from single images without any additional prompt) and PixelLLM~\cite{xu2024cvpr_pixelllm} (which predicts captions from bounding-box location prompts in single images).
For BLIP2~\cite{li2023blip}, we follow CaptionAnything~\cite{wang2023caption} to use the SAM2 mask to crop and mask-out the background.
For PixelLLM~\cite{xu2024cvpr_pixelllm}, we extract the bounding box as the prompt from the SAM2 mask.
These image baselines produce a caption in each frame, and we take a single video-level caption by taking the most common captions for the image caption sequence.

In addition, we create two baselines which closely follow our annotation pipeline: We use SAM2~\cite{ravi2024sam} and UniRef++~\cite{wu2023uniref++} to generate segmentation masks based on a first frame input mask. Then we 
feed these generated segments to our Gemini~\cite{gemini2024pro1.5} pseudo-annotation pipeline (Sec.~\ref{sec:automatic_annotation}).

\input{tables/results_image_and_video_caption}

\para{Video Object Captioning Results.} We finetune VoCap jointly on SAV-Caption-train and VisualGenome~\cite{krishna2017visual}. Tab.~\ref{tab:results_vocap_vs_sam} presents results on the SAV-Caption-val. VoCap significantly outperforms all baselines in captioning at only a minor decrease in segmentation performance compared to SAM2. 
In particular, BLIP2~\cite{li2023blip} and PixelLLM~\cite{xu2024cvpr_pixelllm} yield suboptimal performance, likely since these image-based models do not capture motion. 
More importantly, our results (47.8 CIDEr) surpass applying SAM2 plus Gemini pseudo-labeling (40.5 CIDEr) despite being significantly more efficient (Gemini is much larger than VoCap).
To understand how our model could outperform this strong baseline, we visually inspected the results. We observed that Gemini typically makes mistakes in small objects (presumably due to resolution) and that it has an `actor bias': it sometimes describes a human (hand) or animal which is near the highlighted object. In contrast, since our model actively tracks an object, it always describes the object which it is tracking. Some qualitative examples can be found in Appendix~\ref{sec:qualitative_vocap_vs_sam}. From a more general learning perspective, by training on large amounts of data our model can correct or smooth out some of the noise of the pseudo-labels, which is a commonly observed phenomenon (e.g.~\cite{jia21align,lee2013pseudo,radford2021learning}).

\para{Image Object Captioning.} There are several works on localized image captioning, where the input is an image and a bounding box around an object, and the output is the caption describing the object. Since our model can also consume box prompts, and since images can be interpreted as single-frame videos, we can directly compare to these works. 
We evalate the same VoCap model as before (finetuned jointly on SAV-Caption-train and VisualGenome) and evaluate it on the 5k validation images of VisualGenome~\cite{krishna2017visual} which has human-annotated object captions. 
Again, we report the standard captioning metric, CIDEr~\cite{vedantam2015cider}. 
Results in Tab.~\ref{tab:results_image_caption} show that our method outperforms the state-of-the-art on this task: 150 CIDEr for SCA~\cite{huang2024segment} vs 163 CIDEr for our VoCap model.

\subsection{Video Object Segmentation}\label{sec:results_vos}

In addition to video object captioning, our model can perform both semi-supervised video object segmentation (SS-VOS) and referring expression video object segmentation (RefVOS). In SS-VOS the model input is a video and a ground-truth object mask for the first frame. In RefVOS the input is a video and a textual referring expression of the target object. For both tasks the output is a spatio-temporal masklet throughout the whole video which segments the object in every single frame.

\para{Datasets.}
For SS-VOS we evaluate on the popular YTVOS 2018 dataset~\cite{xu2018youtube} and MOSE~\cite{ding2023mose}.
MOSE was designed to be extra difficult, featuring heavy occlusion and out-of-view motion.
To compare to related works, we do not train on the MOSE training set and instead evaluate in a `zero-shot' setting. Results were obtained using the official test servers on CodaLab.

\input{tables/results_vos}

For RefVOS we evaluate on the popular video referring segmentation datasets RefVOS-YTVOS~\cite{seo2020urvos}, RefVOS-DAVIS~\cite{khoreva2018rvos}, MeVis~\cite{ding2023mevis} and UVO-VLN~\cite{voigtlaender2023connecting}.
For RefVOS-DAVIS~\cite{khoreva2018rvos} we follow UniRef++~\cite{wu2023uniref++} to only use its validation set of 30 videos as a zero-shot evaluation (on average 2 objects per video and with 4 text queries per object). 
The UVO-VLN Video Narrative Grounding (VNG) benchmark provides image descriptions and segmentation masks of labeled noun phrases. To turn a description into a referring expressions we simply mark the target noun with brackets (e.g. `the dog catches the [frisbee]').

Now one problem with referring expressions is that the first frame may not have the clearest view of the object, it could be ambiguous (e.g. for `the bird flying away' there could be three birds where one of them flies away only at the end) or not even visible at the first frame. Such cases are problematic for our model since it will be biases to track keep tracking the object predicted in the first frame.
To overcome this we experimented with the test-time inference method of FindTrack~\cite{cho2025arxiv_findtrack}: We apply VoCap to each frame $t$ independently to produce masks with IoU predictions $iou_t$. We start from the mask and frame with the highest IoU prediction and from there we go both forward and backward in the video to produce a full masklet. Note that with appropriate caching this only requires re-running the mask-decoders twice for each frame, which is less than 10\% extra overhead.
On all RefVOS datasets we report J\&F scores~\cite{perazzi2016davis} (mean of IoU and contour accuracy) averaged on all text queries. Results on RefVOS-YTVOS and MeViS were obtained using the official test servers.

\para{Results.}
Tab.~\ref{tab:results_vos} compares with the state-of-the-art.
Note that SAM2 can only do SS-VOS. ReferFormer~\cite{wu2022language}, SOC~\cite{luo2024soc}, and DsHmp~\cite{he2024cvpr_dshmp} can only do RefVOS. UniRef++~\cite{wu2023uniref++} can handle both tasks, and GLEE~\cite{wu2024general} is a multi-task model which can also perform classical detection and segmentation. However, none of the compared methods can do captioning.

Tab.~\ref{tab:results_vos} shows that on the RefVOS task, if we use FindTrack at test-time, our model outperforms the state of the art on all datasets; by +4.8\% on MeViS, +0.6\% on RefVOS-YTVOS, +0.5\% on RefVOS-DAVIS, and +16.5\% on UVO-VLN. If we do not apply FindTrack~\cite{cho2025arxiv_findtrack}, our model is still best on most datasets except for RefVOS-YTVOS, where GLEE is best. Now GLEE does tracking by detection, which requires making predictions for all frames before it runs an algorithm to merge these per-frame predictions into masklets; it needs to analyze the whole video first. 
In contrast, our model without FindTrack is strictly online which is more applicable in practice but which is a harder task. FindTrack can be used for offline use and gives significant boosts on RefVOS-YTVOS (+0.9\%) and MeViS (+1.1\%).

On SS-VOS our model outperforms the multi-task models GLEE and UniRef++ on semi-supervised visual object segmentation: on YTVOS 2018 our method yields 85.5 $J\&F$ while UniRef++ yields $83.2$ and GLEE yields $80.4$. On the more difficult MOSE dataset the differences are even larger: We obtain 66.3 J\&F while UniRef++ obtains 59.0 and GLEE 56.1.
Overall, we conclude that our model is state-of-the-art on video object segmentation, while it can additionally produce captions.

\input{tables/caption_ablation}

\subsection{Ablation on the effectiveness of SAV-Caption Train and general data mix}

To better understand the importance of our automatically annotated dataset, we ablate the effectiveness of our automatically annotated training set (\refsec{automatic_annotation}).
In particular, we compare our model which was produced by our multi-task training phase (see Tab.~\ref{tab:training_datasets}) to several models which we we train in the same manner but by using increasingly less of our automatically annotated SAV-Caption-train set. However, as SAV is our only video object captioning source we make up for the loss of such data by inverting RefVOS to become a captioning dataset (following~\cite{zhou2023dense}) by considering the query text prompt, which is normally an input, as the output caption for the object. 
In addition, we also show results when doing the full training schedule.
Tab.~\ref{tab:ablation_sav} shows the results.

First of all, the finetuning phase improves results, which is not very surprising.
More importantly, if we use increasingly less SAV-Caption data in multi-task training, SAV-caption performance starts dropping, and completely collapses without any SAV-caption training data.
This demonstrates that our automatic annotation is essential to obtaining good captioning performance on SAV-Caption val.
Interestingly, we observe a significant drop from 68.7\% J\&F to 66.6\% when removing all SAV-caption data. This demonstrates that our model is able able to exploit the synergies between this task and object captioning.

%% file: tables/datasets.tex
\begin{table}[!t]
\vspace{-0mm}
 	\caption{
		\textbf{Datasets used in our multi-task training phase.} Our model can leverage datasets with partial annotations. Note that on our automatically annotated SAV-Caption-train dataset we jointly train on captioning and semi-supervised object segmentation.
	}
        \label{tab:training_datasets}
	\centering
	\begin{tabular}{@{}l@{\quad }l@{\ \ }c@{\ \ }c@{\ \ }c@{}}
		\toprule
		Dataset & Type & Size & Prompt & Output \\
		\midrule
		VisualGenome~\citep{krishna2017visual} & Image & 70K & box & text  \\ 
            RefCOCO~\citep{yu2016modeling} & Image & 24K & text & mask  \\ 
            {\small RefVOS-YTVOS}~\citep{seo2020urvos}\!\! & Video & 3.5K & text & masklets  \\ 
            \savs-train (automatic) & Video & 50K & mask & text, masklets  \\ 
		\bottomrule
	\end{tabular}
	\vspace{-4mm}
\end{table}

%% file: tables/results_image_and_video_caption.tex
\begin{table*}[t] %
\begin{minipage}[t]{0.6\textwidth} %
\centering
\caption{{\bf Video Object Captioning Results.} We significantly outperform strong baselines in video object captioning. \\ $\dagger$ The SAM2 numbers are from our retrained model.}
\label{tab:results_vocap_vs_sam}
\centering
\resizebox{\columnwidth}{!}{  %
\begin{tabular}{@{}lcc@{}}
    \toprule
              & \multicolumn{2}{c}{\bf{SAV-Caption-val (manual)}} \\
    method     &  captioning & segmentation \\
              &  CIDEr  & J\&F \\
    \midrule
    SAM2 $\dagger$ & \na & 75.8 \\
    SAM2 $\dagger$ $\rightarrow$ BLIP2~\cite{li2023blip} & 21.9 & 75.8 \\
    SAM2 $\dagger$ $\rightarrow$ PixelLLM~\cite{xu2024cvpr_pixelllm} & 35.5 & 75.8 \\
    SAM2 $\dagger$ $\rightarrow$ Gemini pseudo-labeling & 40.5 & 75.8 \\
    UniRef++ $\rightarrow$ Gemini pseudo-labeling & 34.3 & 46.9 \\
    \midrule
    VoCap (ours) & 47.8 & 75.5 \\
    \bottomrule
\end{tabular}
}  %
\end{minipage}%
\hfill %
\begin{minipage}[t]{0.38\textwidth} %
\caption{{\bf Results on Localized Image Captioning.} The input is a box, the output a caption. We outperform all existing works.}
\label{tab:results_image_caption}
\resizebox{1.0\columnwidth}{!}{  %
\centering
\begin{tabular}{@{}lc@{}}
    \toprule
              & \bf{Visual Genome} \\
    method     &  captioning - CIDEr \\
    \midrule
    GRiT~\cite{wu2024grit} & 142 \\
    PixelLLM~\cite{xu2024cvpr_pixelllm} & 149 \\
    SCA~\cite{huang2024segment} & 150 \\
    \midrule
    VoCap (ours) & 163 \\
    \bottomrule
\end{tabular}
}  %
\end{minipage}
\end{table*}

%% file: tables/results_vos.tex
\begin{table*}[t!]
\caption{\textbf{State-of-the-art comparison Video Object Segmentation (VOS).} We evaluate both SS-VOS and R-VOS and report official metrics on each dataset. 
`\na` means the model cannot perform the task, and `-` means the paper does not report results.
$\dagger$ SAM2 numbers from our retrained model with eva02 backbone.  %
On MOSE all reported number are zero-shot results; none of the models saw any MOSE training data.
VoCap outperforms the state-of-the-art on all datasets on Referring VOS, and matches the state-of-the-art on SS-VOS.
}
\label{tab:results_vos}
\centering
\resizebox{\textwidth}{!}{
\begin{tabular}{@{}l@{}c@{\quad}c@{\qquad}c@{\quad}c@{\quad}c@{\quad}c@{}}
\toprule
\emph{Task} & \multicolumn{2}{c}{\emph{Semi-Supervised Video Object Segmentation}} & \multicolumn{4}{c}{\emph{Referring Video Object Segmentation}} \\
\emph{Dataset} & \bf{YTVOS 2018} & {\bf MOSE } (zero-shot) & \bf{RefVOS-DAVIS} & \bf{RefVOS-YTVOS}  & {\bf MeViS}  & \bf{UVO-VLN} \\
metric & $\mathcal{G}$ & J\&F  &  J\&F  & J\&F & J\&F \\
\cmidrule(r){1-1}
\cmidrule(r){2-3}
\cmidrule(r){4-7}
SAM2~\cite{ravi2024sam}$\dagger$ & {\bf 85.0} & {\bf 66.4} & \na & \na & \na & \na \\
Point-VOS~\cite{zulfikar2024cvpr_pointvos} & 73.7 & - & - & - & - & 52.8 \\
ReferFormer~\cite{wu2022language} & \na & \na & 61.1 & 64.9 & - & 46.4 \\
SOC~\cite{luo2024soc} & \na & \na & 67.2 & 67.3 & - & - \\
DsHmp~\cite{he2024cvpr_dshmp} & \na & \na & 64.9 & 67.1 & 46.4 & - \\
FindTrack~\cite{cho2025arxiv_findtrack} & \na & \na & 74.2 & 70.3 & 48.2 & - \\
UniRef++~\cite{wu2023uniref++} & 83.2 & 59.0 & 67.2 & 67.4 & - & - \\
GLEE~\cite{wu2024general} & 80.4 & 56.1 & - & 70.6 & - & - \\
\cmidrule(r){1-1}
\cmidrule(r){2-3}
\cmidrule(r){4-7}
VoCap (ours) & {\bf 85.0} & 66.3 & {\bf 75.1} & 70.3 & 51.9 & 62.2 \\
VoCap + FindTrack (ours) & \na & \na & 74.7 & {\bf 71.2} & {\bf 53.0} & 62.7 \\
\bottomrule
\end{tabular}
}

\end{table*}

%% file: tables/caption_ablation.tex
\begin{table}[!t]
\caption{
    \textbf{Effectiveness of the \savs training data.}
    We show results for both video object captioning (mask-to-text) and referring object segmentation (text-to-masklet).
    Our \savs training set improves both tasks.
}
\label{tab:ablation_sav}
\centering
\begin{tabular}{@{\ }l@{\quad}c@{\ \ }c@{\ \ }c@{}}
    \toprule
              &  \multicolumn{2}{c}{{\bf SAV-Caption-val (manual)}}  &  {\bf RefVOS-YTVOS} \\
              & Captioning & SS-VOS & RefVOS \\
              & CIDEr & J\&F & J\&F \\
    \cmidrule(r){2-3}
    \cmidrule(r){4-4}
    Full training (phase \emph{i}, \emph{ii}, and \emph{iii}) & 47.8 & 75.5 & 70.3 \\
    Pre-training and Multitask training (phase \emph{i} and \emph{ii}) & 44.8 & 75.1 & 68.7 \\ %
    $\hookrightarrow$ 50\% of \savs-train & 42.1 & 74.9 & 69.2 \\ %
    $\hookrightarrow$ 25\% of \savs-train & 42.1 & 72.0 & 68.2 \\ %
    $\hookrightarrow$ 0\% of \savs-train & 27.4 & 57.7 & 66.6 \\ %
    \bottomrule
\end{tabular}
\vspace{-0mm}
\end{table}

%% file: 6_conclusion.tex
\section{Conclusion}
\label{sec:conclusion}

We proposed a video object segmentation and captioning model that takes either a box, mask or text prompt as input. 
We manually collected evaluation data for this task, and proposed an automatic annotation pipeline to curate training data.
VoCap trained on our \savs dataset together with diverse existing datasets 
outperforms the state-of-the-art on referring expression video object segmentation and reaches top-tier performance on 
our captioning task and semi-supervised video segmentation.
We hope our model and datasets provide a foundation for fine-grained spatio-temporal video understanding, and encourages more work in this direction.

\par  %

%% file: appendix_arxiv.tex
\section{Details on Dataset}

\subsection{Quality of SAV-Caption-train}

We performed a quantitative evaluation on the quality of the SAV-Caption-train set by having the authors examine captions of 50 randomly selected objects from 50 different videos. They verified separately whether each of the elements used from our structured prompt was correct or not: the object category, its properties, and what the object does (e.g. motion or action). Furthermore, we counted how many properties and actions were obvious yet not generated in the caption. Results are in Tab.~\ref{tab:sav_quality}.

\begin{table}[hbt]
    \centering
    \caption{Quantitative analysis of the quality of SAV-Caption-train on 50 objects in 50 different videos. \# evaluated means the number of respectively categories, properties, and motion/actions we evaluated.}
    \begin{tabular}{lcccc}
         \toprule
         & correct & incorrect (hallucinations) & \# evaluated & \# missing aspects \\
         \midrule
         object category & 88.0\% & 12.0\% & 50 & - \\
         object properties & 87.6\% & 12.4\% & 105 & 7 \\
         object motion / action & 85.5\% & 15.5\% & 62 & 5 \\
         \bottomrule
    \end{tabular}
    \label{tab:sav_quality}
\end{table}

The object category was correct in 88.0\% of the cases. When an object was incorrect it was either subtle (e.g. \emph{sock} instead of \emph{shoe}) or it was a piece of clothing worn by a human and the human was captioned instead. Properties are also correct in 87.6\% of the cases. Many mistakes were subtle color differences due to lighting conditions. When the human was described instead of their clothing worn, we counted these properties as incorrect (even if they were correct for the human).
There were a few properties noticeably absent, mostly because of the context of other mentioned properties. For example, one caption mentioning a white striped sweater, whereas the sweater was \emph{blue}-white striped, which conveys a quite different appearance of the sweater.
The object's motion/action was correct in 85.5\% of the cases. Most mistakes were subtle differences between standing still or driving / walking slowly. Similarly as before, when the motion/action was of the wrong category (person instead of sweater), we counted this as incorrect. In 7 instances we found an action to be clearly missing. This was usually when the object did multiple things sequentially (e.g. a person is first standing, then walking away) where one of them was missing. In another there was a parrot which was correctly identified to be laying on the floor, but they did this to scratch their head on the floor; a crucial aspect to understand its behavior.

To conclude, while there is some noise in the automatically generated data we consider it to be of decent quality. Moreover, in our main paper we clearly demonstrate the usefulness of this data for training captioning models.

\subsection{Text prompt to generate \savstrain}

We use the following prompt together with our vision prompts to generate the pseudo-labels of our training set.

{\footnotesize
\noindent \texttt{
\!\!\!\!\!\! Describe the subject in the red contour in the following video.
If the subject is a part of an object, please describe this part instead of the whole object.
Please DO NOT DESCRIBE anything in the blurred background outside the red contour.
First determine the subject's category ({CATEGORY}), properties ({PROPERTIES}), action ({ACTION}), and then give a description in ONE sentence ({DESCRIPTION}) including category, properties, and action, etc..
Please use this FORMAT: 'The video shows a {CATEGORY}. The subject's properties are {PROPERTIES}. The subject's action is {ACTION}. {DESCRIPTION}.'.
The {DESCRIPTION} starts with 'A/ An {CATEGORY}' or 'A/ An {PROPERTIES} {CATEGORY}' if it is grammarly more proper to put the properties before the category.
The category, properties, motion and the descriptions should be consistent.
{PROPERTIES} should be about the objects appearance (color, texture, size, material, shape), what it is wearing or a functional property (e.g. fast, sharp). Please always include interesting or unexpected properties.
If there are multiple actions happening sequentially, connect them with 'then', but do not include more than 3 actions.
For static objects or parts, just say the {ACTION} is 'static' and it is OK to not include {ACTION} in {DESCRIPTION}.
Please DO NOT mention the red contour in the description.
If the subject is a person, please avoid describing the person's skin color and describe the person's clothes color instead.
You only need to describe the details that you are certain about.
If you cannot perform the task or you are very uncertain, please say `I cannot perform the task for this video.'.
}
}

\section{SAM2 baseline details}\label{sec:sam2_replication}

The original SAM2~\cite{ravi2024sam} was trained on private datasets in addition to the publicly-released SAV training and validation set.
The publicly-released SAM2 training code\footnote{\href{https://github.com/facebookresearch/sam2}{https://github.com/facebookresearch/sam2}} includes finetuning pipeline on MOSE dataset~\cite{ding2023mose}, but does not include the main training loop.
Therefore, before adapting SAM2 in our use case, we attempt to reproduce SAM2 training in our framework in Jax~\cite{jax2018github, dehghani2022scenic}.
We also repleace the MAE-pretrained backbone Hiera~\cite{ryali2023hiera} with a more vision-langauge native backbone Eva02~\cite{EVA02}.
When using Eva02, we reduce the input resolution from the original 1024 to 512 to fit our hardware, and verified minimal performance drop compare to the official SAM2 with Hiera-T and 1024 input size.
We do not use the SA1B dataset~\cite{kirillov2023segment}
for pretraining as we did not find it helpful in our target datasets.
We adapt the training hyper-parameters in Table 12 (b) of the SAM2 paper, which we summarize in~\reftbl{hyperparams-sam2}.

\begin{table}[h!]
\small
\centering
\caption{\textbf{Results of our reproduced SAM2}. We use a vision-language native backbone Eva02~\cite{EVA02} with a smaller input size (512 vs. 1024), and show the performance matches the original SAM2.}
\lbltbl{sam2-reproduce}
\begin{tabular}{lcc|cc}
    \toprule
           & backbone & resolution  &  MOSE-dev & SAV-val \\
    \midrule
    Official SAM2 & Hiera-L & 1024 & 77.9 & 77.9 \\
    Official SAM2 & Hiera-T & 512 & 75.3 & 75.2 \\
    Our reproduction & Hiera-T & 512 & 76.9 & 74.8 \\
    Our reproduction & EVA02-L & 512 & 75.7 & 75.8 \\
    \bottomrule
\end{tabular}
\end{table}

\begin{table}[t]
\centering
\caption{\textbf{Hyperparameters of our reproduce of SAM2 as our pre-training.} We follow SAM2~\cite{ravi2024sam} for most hyperparameters.}
\lbltbl{hyperparams-sam2}
\begin{tabular}{@{}c|@{}c@{}}
config  & value \\
\hline
data & SA-V, YTVOS, DAVIS \\
data-ratio & 49.5: 9.4: 1.3 \\
steps & 300k \\
backbone & Hiera-T / Eva02 \\
resolution & 896 (Hiera-T) / 512 (Eva02) \\
optimizer & AdamW \\
optimizer momentum & {$\beta_1, \beta_2{=}0.9, 0.999$} \\ 
gradient clipping & type: $\ell_2$, max: {0.1} \\
weight decay & 0.05 \\
learning rate (lr) & img. enc.: 4e-5, other: 4.0e-4 \\
lr schedule & cosine\\
warmup & linear, 1k iters \\ 
layer-wise decay & 0.8 \\
augmentation & hflip, crop and square resize to 512\\
batch size & 64 \\
drop path & 0.1 (Hiera-T) / 0.4 (Eva02) \\
mask losses (weight) & focal (20), dice (1)\\
IoU loss (weight) & $\ell_1$ (1)\\
occlusion loss (weight) & cross-entropy (1) \\
num frames & 8 \\
max. masks per frame. & 2 \\
\end{tabular}
\end{table}

As a result, our reproduced SAM2 with Eva02~\cite{EVA02} and a smaller input size trained on public data closely matches the official released model, as shown~\reftbl{sam2-reproduce}.

\section{Qualitative examples: VoCap vs SAM2 + Gemini Pseudo captioning}\label{sec:qualitative_vocap_vs_sam}

Tab.~\ref{tab:results_vocap_vs_sam} demonstrates that VoCap outperforms the strong baseline of applying SAM2 followed by Gemini pseudo-labeling. Figure~\ref{fig:vocap_better_than_baseline} illustrates why: The Gemini captions are often wrong for small objects (both examples). Furthermore, when a human (hand) or animal is near the object, Gemini 1.5 Pro sometimes describes this actor instead of the highlighted object (example on the left in Fig.~\ref{fig:vocap_better_than_baseline}).

\input{figures/qualitative_results_vocap_better_than_baseline}

\section{Training hyper-parameters}

We include the full hyper-parameters used in our multi-task training phase in~\reftbl{hyperparams-main}

\begin{table}[t]
\centering
\begin{tabular}{@{}c|@{}c@{}}
config  & value \\
\hline
\multirow{2}{*}{data} & \savs, RefVOS-YTVOS, \\
 & RefCOCO, VisualGenome \\
data-ratio & 2: 1: 1: 0.5 \\
steps & 240k \\
backbone & Eva02 \\
resolution & 512 \\
optimizer & AdamW \\
optimizer momentum & {$\beta_1, \beta_2{=}0.9, 0.999$} \\ 
gradient clipping & type: $\ell_2$, max: {0.1} \\
weight decay & 0.05 \\
learning rate (lr) & 5e-5 \\
lr schedule & cosine\\
warmup & linear, 1k iters \\ 
layer-wise decay & 0.8 \\
augmentation & crop and square resize to 512\\
batch size & 32 \\
drop path & 0.4 \\
mask losses (weight) & focal (20), dice (1)\\
IoU loss (weight) & $\ell_1$ (1)\\
occlusion loss (weight) & cross-entropy (1) \\
caption loss (weight) & cross-entropy (1) \\
caption loss label smooth & 0.1 \\
num frames & 8 \\
max. masks per frame. & image: 32, video: 2 \\
\end{tabular}
\caption{\textbf{Hyperparameters of multi-task VoCap training.} We continue train on datasets with both text and mask annotations on both images and videos, with a cross-entropy caption-loss.}
\lbltbl{hyperparams-main}
\end{table}

%% file: figures/qualitative_results_vocap_better_than_baseline.tex
\begin{figure}[t]
    \captionsetup[subfigure]{labelformat=empty}
    \centering
    \begin{subfigure}{0.4\textwidth}
    \includegraphics[width=\linewidth]{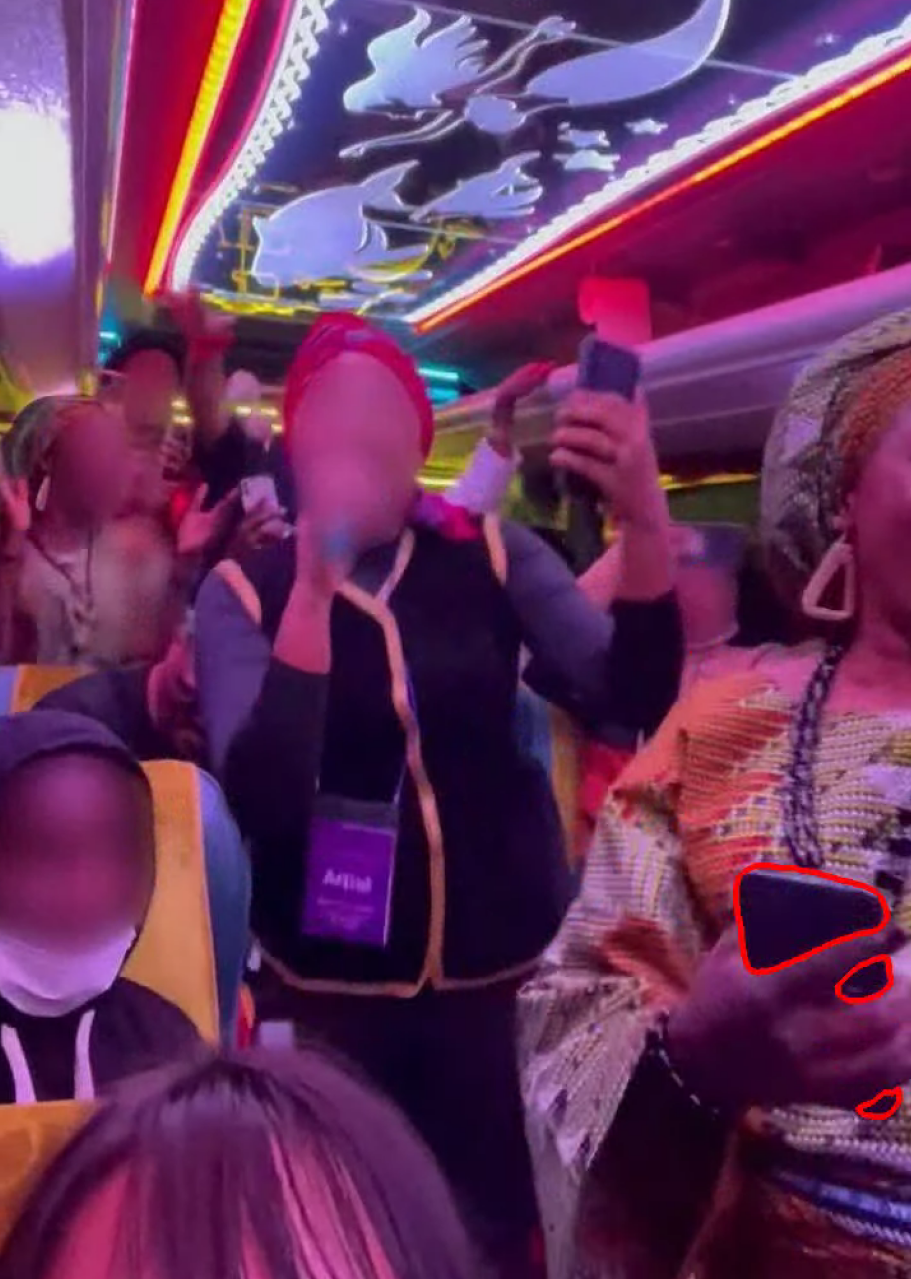}
    \caption{{\bf GT:} A mobile phone, held in the hand of the person who is dancing.\\{\bf SAM2+Gemini:} A hand is holding something.\\{\bf VoCap:} A small, rectangular, dark phone is being held.}
    \end{subfigure}
    \hspace{.5cm}
    \hspace{.5cm}
    \begin{subfigure}{0.4\textwidth}
    \includegraphics[width=\linewidth]{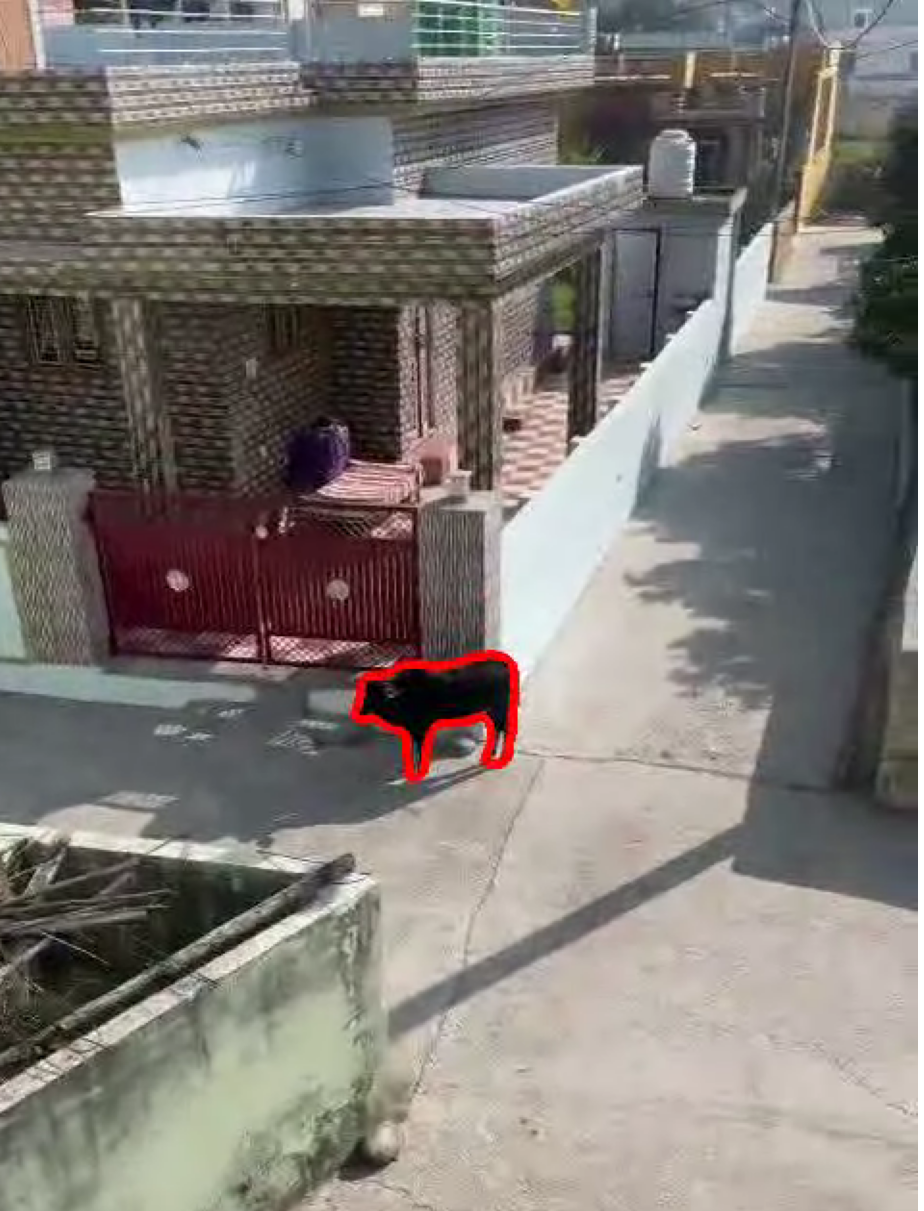}
    \caption{{\bf GT:} A black cow is walking on the road near a house.\\{\bf SAM2+Gemini:} A small black dog is walking on the street.\\{\bf VoCap:} A black cow is walking.}
    \end{subfigure}
    \caption{Qualitative examples which illustrate where VoCap succeesd where SAM2+Gemini pseudo labeling does not. This typically happens in cases with small objects (both examples) and when an actor is nearby (the phone is held by a human hand).}
    \label{fig:vocap_better_than_baseline}
\end{figure}

%% file: neurips_2025.bbl
\begin{thebibliography}{110}
\providecommand{\natexlab}[1]{#1}
\providecommand{\url}[1]{\texttt{#1}}
\expandafter\ifx\csname urlstyle\endcsname\relax
  \providecommand{\doi}[1]{doi: #1}\else
  \providecommand{\doi}{doi: \begingroup \urlstyle{rm}\Url}\fi

\bibitem[Alayrac et~al.(2022)Alayrac, Donahue, Luc, Miech, Barr, Hasson, Lenc,
  Mensch, Millican, Reynolds, et~al.]{alayrac2022flamingo}
Jean-Baptiste Alayrac, Jeff Donahue, Pauline Luc, Antoine Miech, Iain Barr,
  Yana Hasson, Karel Lenc, Arthur Mensch, Katherine Millican, Malcolm Reynolds,
  et~al.
\newblock Flamingo: a visual language model for few-shot learning.
\newblock \emph{NeurIPS}, 2022.

\bibitem[Bain et~al.(2021)Bain, Nagrani, Varol, and Zisserman]{bain2021frozen}
Max Bain, Arsha Nagrani, G{\"u}l Varol, and Andrew Zisserman.
\newblock Frozen in time: A joint video and image encoder for end-to-end
  retrieval.
\newblock In \emph{CVPR}, 2021.

\bibitem[Beery et~al.(2020)Beery, Wu, Rathod, Votel, and
  Huang]{beery2020context}
Sara Beery, Guanhang Wu, Vivek Rathod, Ronny Votel, and Jonathan Huang.
\newblock Context r-cnn: Long term temporal context for per-camera object
  detection.
\newblock In \emph{CVPR}, 2020.

\bibitem[Bradbury et~al.(2018)Bradbury, Frostig, Hawkins, Johnson, Leary,
  Maclaurin, Necula, Paszke, Vander{P}las, Wanderman-{M}ilne, and
  Zhang]{jax2018github}
James Bradbury, Roy Frostig, Peter Hawkins, Matthew~James Johnson, Chris Leary,
  Dougal Maclaurin, George Necula, Adam Paszke, Jake Vander{P}las, Skye
  Wanderman-{M}ilne, and Qiao Zhang.
\newblock {JAX}: composable transformations of {P}ython+{N}um{P}y programs,
  2018.

\bibitem[Caelles et~al.(2019)Caelles, Pont-Tuset, Perazzi, Montes, Maninis, and
  {Van Gool}]{caelles2019davis}
Sergi Caelles, Jordi Pont-Tuset, Federico Perazzi, Alberto Montes,
  Kevis-Kokitsi Maninis, and Luc {Van Gool}.
\newblock The 2019 davis challenge on vos: Unsupervised multi-object
  segmentation.
\newblock \emph{arXiv:1905.00737}, 2019.

\bibitem[Caesar et~al.(2020)Caesar, Bankiti, Lang, Vora, Liong, Xu, Krishnan,
  Pan, Baldan, and Beijbom]{caesar2020nuscenes}
Holger Caesar, Varun Bankiti, Alex~H Lang, Sourabh Vora, Venice~Erin Liong,
  Qiang Xu, Anush Krishnan, Yu Pan, Giancarlo Baldan, and Oscar Beijbom.
\newblock nuscenes: A multimodal dataset for autonomous driving.
\newblock In \emph{CVPR}, 2020.

\bibitem[Chai et~al.(2023)Chai, Guo, Wang, and Lu]{chai2023stablevideo}
Wenhao Chai, Xun Guo, Gaoang Wang, and Yan Lu.
\newblock Stablevideo: Text-driven consistency-aware diffusion video editing.
\newblock In \emph{CVPR}, 2023.

\bibitem[Chen et~al.(2022)Chen, Wang, Changpinyo, Piergiovanni, Padlewski,
  Salz, Goodman, Grycner, Mustafa, Beyer, et~al.]{chen2022pali}
Xi Chen, Xiao Wang, Soravit Changpinyo, AJ Piergiovanni, Piotr Padlewski,
  Daniel Salz, Sebastian Goodman, Adam Grycner, Basil Mustafa, Lucas Beyer,
  et~al.
\newblock Pali: A jointly-scaled multilingual language-image model.
\newblock \emph{arXiv:2209.06794}, 2022.

\bibitem[Cheng et~al.(2021)Cheng, Schwing, and Kirillov]{cheng2021per}
Bowen Cheng, Alex Schwing, and Alexander Kirillov.
\newblock Per-pixel classification is not all you need for semantic
  segmentation.
\newblock \emph{NeurIPS}, 2021.

\bibitem[Cheng and Schwing(2022)]{cheng2022xmem}
Ho~Kei Cheng and Alexander~G Schwing.
\newblock Xmem: Long-term video object segmentation with an atkinson-shiffrin
  memory model.
\newblock In \emph{ECCV}, 2022.

\bibitem[Cheng et~al.(2024)Cheng, Oh, Price, Lee, and
  Schwing]{cheng2024putting}
Ho~Kei Cheng, Seoung~Wug Oh, Brian Price, Joon-Young Lee, and Alexander
  Schwing.
\newblock Putting the object back into video object segmentation.
\newblock In \emph{CVPR}, 2024.

\bibitem[Cheng et~al.(2023)Cheng, Li, Xu, Li, Yang, Wang, and
  Yang]{cheng2023segment}
Yangming Cheng, Liulei Li, Yuanyou Xu, Xiaodi Li, Zongxin Yang, Wenguan Wang,
  and Yi Yang.
\newblock Segment and track anything.
\newblock \emph{arXiv:2305.06558}, 2023.

\bibitem[Cho et~al.(2025)Cho, Lee, Lee, Lee, and Lee]{cho2025arxiv_findtrack}
Suhwan Cho, Seunghoon Lee, Minhyeok Lee, Jungho Lee, and Sangyoun Lee.
\newblock Find first, track next: Decoupling identification and propagation in
  referring video object segmentation.
\newblock In \emph{ArXiv}, 2025.

\bibitem[Choudhuri et~al.(2024)Choudhuri, Chowdhary, and
  Schwing]{choudhuri2024ow}
Anwesa Choudhuri, Girish Chowdhary, and Alexander~G Schwing.
\newblock Ow-viscaptor: Abstractors for open-world video instance segmentation
  and captioning.
\newblock \emph{NeurIPS}, 2024.

\bibitem[Dehghani et~al.(2022)Dehghani, Gritsenko, Arnab, Minderer, and
  Tay]{dehghani2022scenic}
Mostafa Dehghani, Alexey Gritsenko, Anurag Arnab, Matthias Minderer, and Yi
  Tay.
\newblock Scenic: A jax library for computer vision research and beyond.
\newblock In \emph{Proceedings of the IEEE/CVF conference on computer vision
  and pattern recognition}, pages 21393--21398, 2022.

\bibitem[Deng et~al.(2024)Deng, Wu, Zeng, and Qin]{deng2024memsam}
Xiaolong Deng, Huisi Wu, Runhao Zeng, and Jing Qin.
\newblock Memsam: Taming segment anything model for echocardiography video
  segmentation.
\newblock In \emph{CVPR}, 2024.

\bibitem[Devlin et~al.(2019)Devlin, Chang, Lee, and Toutanova]{devlin2018bert}
Jacob Devlin, Ming-Wei Chang, Kenton Lee, and Kristina Toutanova.
\newblock Bert: Pre-training of deep bidirectional transformers for language
  understanding.
\newblock In \emph{NAACL}, 2019.

\bibitem[Ding et~al.(2023{\natexlab{a}})Ding, Liu, He, Jiang, and
  Loy]{ding2023mevis}
Henghui Ding, Chang Liu, Shuting He, Xudong Jiang, and Chen~Change Loy.
\newblock Mevis: A large-scale benchmark for video segmentation with motion
  expressions.
\newblock In \emph{ICCV}, 2023{\natexlab{a}}.

\bibitem[Ding et~al.(2023{\natexlab{b}})Ding, Liu, He, Jiang, Torr, and
  Bai]{ding2023mose}
Henghui Ding, Chang Liu, Shuting He, Xudong Jiang, Philip~HS Torr, and Song
  Bai.
\newblock {MOSE}: A new dataset for video object segmentation in complex
  scenes.
\newblock In \emph{ICCV}, 2023{\natexlab{b}}.

\bibitem[Dosovitskiy(2021)]{dosovitskiy2020image}
Alexey Dosovitskiy.
\newblock An image is worth 16x16 words: Transformers for image recognition at
  scale.
\newblock \emph{ICLR}, 2021.

\bibitem[Fang et~al.(2023)Fang, Sun, Wang, Huang, Wang, and Cao]{EVA02}
Yuxin Fang, Quan Sun, Xinggang Wang, Tiejun Huang, Xinlong Wang, and Yue Cao.
\newblock Eva-02: A visual representation for neon genesis.
\newblock \emph{arXiv:2303.11331}, 2023.

\bibitem[Gemini~Team(2024)]{gemini2024pro1.5}
Google Gemini~Team.
\newblock Gemini 1.5: Unlocking multimodal understanding across millions of
  tokens of context.
\newblock \emph{arXiv:2403.05530}, 2024.

\bibitem[Guo et~al.(2024)Guo, Zhao, Gao, Wu, He, Zhang, Xiao, and
  Zhang]{guo2024videosam}
Pinxue Guo, Zixu Zhao, Jianxiong Gao, Chongruo Wu, Tong He, Zheng Zhang,
  Tianjun Xiao, and Wenqiang Zhang.
\newblock Videosam: Open-world video segmentation.
\newblock \emph{arXiv:2410.08781}, 2024.

\bibitem[He et~al.(2022)He, Chen, Xie, Li, Doll{\'a}r, and
  Girshick]{he2022masked}
Kaiming He, Xinlei Chen, Saining Xie, Yanghao Li, Piotr Doll{\'a}r, and Ross
  Girshick.
\newblock Masked autoencoders are scalable vision learners.
\newblock In \emph{CVPR}, 2022.

\bibitem[He and Ding(2024)]{he2024cvpr_dshmp}
Shuting He and Henghui Ding.
\newblock Decoupling static and hierarchical motion perception for referring
  video segmentation.
\newblock In \emph{Proceedings of the IEEE/CVF Conference on Computer Vision
  and Pattern Recognition}, pages 13332--13341, 2024.

\bibitem[Hu et~al.(2024)Hu, Chan, Su, Chen, Li, Sohn, Zhao, Ben, Gong, Cohen,
  et~al.]{hu2024instruct}
Hexiang Hu, Kelvin~CK Chan, Yu-Chuan Su, Wenhu Chen, Yandong Li, Kihyuk Sohn,
  Yang Zhao, Xue Ben, Boqing Gong, William Cohen, et~al.
\newblock Instruct-imagen: Image generation with multi-modal instruction.
\newblock In \emph{CVPR}, 2024.

\bibitem[Huang et~al.(2024)Huang, Wang, Tang, Zhang, Hu, Lu, Wang, and
  Liu]{huang2024segment}
Xiaoke Huang, Jianfeng Wang, Yansong Tang, Zheng Zhang, Han Hu, Jiwen Lu,
  Lijuan Wang, and Zicheng Liu.
\newblock Segment and caption anything.
\newblock In \emph{CVPR}, 2024.

\bibitem[Iashin and Rahtu(2020)]{iashin2020bmvc_vidcapbimodal}
Vladimir Iashin and Esa Rahtu.
\newblock A better use of audio-visual cues: Dense video captioning with
  bi-modal transformer.
\newblock In \emph{BMVC}, 2020.

\bibitem[Jaegle et~al.(2021)Jaegle, Gimeno, Brock, Vinyals, Zisserman, and
  Carreira]{jaegle2021perceiver}
Andrew Jaegle, Felix Gimeno, Andy Brock, Oriol Vinyals, Andrew Zisserman, and
  Joao Carreira.
\newblock Perceiver: General perception with iterative attention.
\newblock In \emph{ICML}. PMLR, 2021.

\bibitem[Jia et~al.(2021)Jia, Yang, Xia, Chen, Parekh, Pham, Le, Sung, Li, and
  Duerig]{jia21align}
Chao Jia, Yinfei Yang, Ye Xia, Yi-Ting Chen, Zarana Parekh, Hieu Pham, Quoc Le,
  Yun-Hsuan Sung, Zhen Li, and Tom Duerig.
\newblock Scaling up visual and vision-language representation learning with
  noisy text supervision.
\newblock In \emph{ICML}, 2021.

\bibitem[Johnson et~al.(2016)Johnson, Karpathy, and
  Fei-Fei]{johnson2016cvpr_densecap}
Justin Johnson, Andrej Karpathy, and Li Fei-Fei.
\newblock Densecap: Fully convolutional localization networks for dense
  captioning.
\newblock In \emph{CVPR}, 2016.

\bibitem[Kanani et~al.(2021)Kanani, Saha, and
  Bhattacharyya]{kanani2021ijcnn_videodescriptions}
Chandresh~S. Kanani, Sriparna Saha, and Pushpak Bhattacharyya.
\newblock Global object proposals for improving multi-sentence video
  descriptions.
\newblock In \emph{IJCNN}, 2021.

\bibitem[Khoreva et~al.(2018)Khoreva, Rohrbach, and Schiele]{khoreva2018rvos}
Anna Khoreva, Anna Rohrbach, and Bernt Schiele.
\newblock Video object segmentation with language referring expressions.
\newblock In \emph{ACCV}, 2018.

\bibitem[Kim et~al.(2020)Kim, Woo, Lee, and Kweon]{kim2020cvpr_panopticvideo}
Dahun Kim, Sanghyun Woo, Joon-Young Lee, and In~So Kweon.
\newblock Video panoptic segmentation.
\newblock In \emph{CVPR}, 2020.

\bibitem[Kirillov et~al.(2023)Kirillov, Mintun, Ravi, Mao, Rolland, Gustafson,
  Xiao, Whitehead, Berg, Lo, et~al.]{kirillov2023segment}
Alexander Kirillov, Eric Mintun, Nikhila Ravi, Hanzi Mao, Chloe Rolland, Laura
  Gustafson, Tete Xiao, Spencer Whitehead, Alexander~C Berg, Wan-Yen Lo, et~al.
\newblock Segment anything.
\newblock In \emph{ICCV}, 2023.

\bibitem[Krishna et~al.(2017{\natexlab{a}})Krishna, Hata, Ren, Fei-Fei, and
  Carlos~Niebles]{krishna2017dense}
Ranjay Krishna, Kenji Hata, Frederic Ren, Li Fei-Fei, and Juan Carlos~Niebles.
\newblock Dense-captioning events in videos.
\newblock In \emph{ICCV}, 2017{\natexlab{a}}.

\bibitem[Krishna et~al.(2017{\natexlab{b}})Krishna, Zhu, Groth, Johnson, Hata,
  Kravitz, Chen, Kalantidis, Li, Shamma, et~al.]{krishna2017visual}
Ranjay Krishna, Yuke Zhu, Oliver Groth, Justin Johnson, Kenji Hata, Joshua
  Kravitz, Stephanie Chen, Yannis Kalantidis, Li-Jia Li, David~A Shamma, et~al.
\newblock Visual genome: Connecting language and vision using crowdsourced
  dense image annotations.
\newblock \emph{IJCV}, 2017{\natexlab{b}}.

\bibitem[Lan et~al.(2023)Lan, Rong, and Zhang]{lan2023referring}
Meng Lan, Fu Rong, and Lefei Zhang.
\newblock Referring video object segmentation with inter-frame interaction and
  cross-modal correlation.
\newblock \emph{arXiv:2307.00536}, 2023.

\bibitem[Lee(2013)]{lee2013pseudo}
Dong-Hyun Lee.
\newblock Pseudo-label: The simple and efficient semi-supervised learning
  method for deep neural networks.
\newblock In \emph{ICML Workshop}, 2013.

\bibitem[Li et~al.(2023)Li, Li, Savarese, and Hoi]{li2023blip}
Junnan Li, Dongxu Li, Silvio Savarese, and Steven Hoi.
\newblock Blip-2: Bootstrapping language-image pre-training with frozen image
  encoders and large language models.
\newblock In \emph{ICML}, 2023.

\bibitem[Li et~al.(2019)Li, Jiang, and Han]{li2019aaai_densecap}
Xiangyang Li, Shuqiang Jiang, and Jungong Han.
\newblock Learning object context for dense captioning.
\newblock In \emph{AAAI}, 2019.

\bibitem[Li et~al.(2022)Li, Mao, Girshick, and He]{li2022exploring}
Yanghao Li, Hanzi Mao, Ross Girshick, and Kaiming He.
\newblock Exploring plain vision transformer backbones for object detection.
\newblock In \emph{ECCV}, 2022.

\bibitem[Li et~al.(2024)Li, Li, Wang, Ma, Yao, Dong, Fan, and
  Zhang]{li2024eccv_bensmot}
Yunhao Li, Qin Li, Hao Wang, Xue Ma, Jiali Yao, Shaohua Dong, Heng Fan, and
  Libo Zhang.
\newblock Beyond mot: Semantic multi-object tracking.
\newblock In \emph{ECCV}, 2024.

\bibitem[Lin et~al.(2024)Lin, Wei, An, Gao, Zou, Luo, Huang, Zhang, and
  Li]{lin2024draw}
Weifeng Lin, Xinyu Wei, Ruichuan An, Peng Gao, Bocheng Zou, Yulin Luo, Siyuan
  Huang, Shanghang Zhang, and Hongsheng Li.
\newblock Draw-and-understand: Leveraging visual prompts to enable mllms to
  comprehend what you want.
\newblock \emph{arXiv:2403.20271}, 2024.

\bibitem[Liu et~al.(2023)Liu, Li, Wu, and Lee]{liu2023llava}
Haotian Liu, Chunyuan Li, Qingyang Wu, and Yong~Jae Lee.
\newblock Visual instruction tuning.
\newblock In \emph{NeurIPS}, 2023.

\bibitem[L{\"u}ddecke and Ecker(2022)]{luddecke2022cvpr_clipseg}
Timo L{\"u}ddecke and Alexander Ecker.
\newblock Image segmentation using text and image prompts.
\newblock In \emph{CVPR}, 2022.

\bibitem[Luo et~al.(2024)Luo, Xiao, Liu, Li, Wang, Tang, Li, and
  Yang]{luo2024soc}
Zhuoyan Luo, Yicheng Xiao, Yong Liu, Shuyan Li, Yitong Wang, Yansong Tang, Xiu
  Li, and Yujiu Yang.
\newblock Soc: Semantic-assisted object cluster for referring video object
  segmentation.
\newblock \emph{NeurIPS}, 2024.

\bibitem[Mao et~al.(2016)Mao, Huang, Toshev, Camburu, Yuille, and
  Murphy]{mao2016generation}
Junhua Mao, Jonathan Huang, Alexander Toshev, Oana Camburu, Alan~L Yuille, and
  Kevin Murphy.
\newblock Generation and comprehension of unambiguous object descriptions.
\newblock In \emph{CVPR}, 2016.

\bibitem[Minderer et~al.(2023)Minderer, Gritsenko, and
  Houlsby]{minderer2023neurips_owlvit2}
Matthias Minderer, Alexey Gritsenko, and Neil Houlsby.
\newblock Scaling open-vocabulary object detection.
\newblock \emph{NeurIPS}, 2023.

\bibitem[Nasiriany et~al.(2024)Nasiriany, Xia, Yu, Xiao, Liang, Dasgupta, Xie,
  Driess, Wahid, Xu, et~al.]{nasiriany2024icml_pivot}
Soroush Nasiriany, Fei Xia, Wenhao Yu, Ted Xiao, Jacky Liang, Ishita Dasgupta,
  Annie Xie, Danny Driess, Ayzaan Wahid, Zhuo Xu, et~al.
\newblock Pivot: Iterative visual prompting elicits actionable knowledge for
  vlms.
\newblock \emph{ICML}, 2024.

\bibitem[OpenAI(2023,)]{gpt4vision}
OpenAI.
\newblock Gpt-4v(ision) technical work and authors.
\newblock \url{https://cdn.openai.com/contributions/gpt-4v.pdf}, 2023,.

\bibitem[Peng et~al.(2023)Peng, Wang, Dong, Hao, Huang, Ma, and
  Wei]{peng2023kosmos}
Zhiliang Peng, Wenhui Wang, Li Dong, Yaru Hao, Shaohan Huang, Shuming Ma, and
  Furu Wei.
\newblock Kosmos-2: Grounding multimodal large language models to the world.
\newblock \emph{arXiv:2306.14824}, 2023.

\bibitem[Perazzi et~al.(2016)Perazzi, Pont-Tuset, McWilliams, {Van Gool},
  Gross, and Sorkine-Hornung]{perazzi2016davis}
F. Perazzi, J. Pont-Tuset, B. McWilliams, L. {Van Gool}, M. Gross, and A.
  Sorkine-Hornung.
\newblock A benchmark dataset and evaluation methodology for video object
  segmentation.
\newblock In \emph{CVPR}, 2016.

\bibitem[Plummer et~al.(2015)Plummer, Wang, Cervantes, Caicedo, Hockenmaier,
  and Lazebnik]{plummer2015cvpr_flickr30k}
Bryan~A Plummer, Liwei Wang, Chris~M Cervantes, Juan~C Caicedo, Julia
  Hockenmaier, and Svetlana Lazebnik.
\newblock Flickr30k entities: Collecting region-to-phrase correspondences for
  richer image-to-sentence models.
\newblock In \emph{CVPR}, 2015.

\bibitem[Pont-Tuset et~al.(2020)Pont-Tuset, Uijlings, Changpinyo, Soricut, and
  Ferrari]{ponttuset2020eccv_locnar}
Jordi Pont-Tuset, Jasper Uijlings, Soravit Changpinyo, Radu Soricut, and
  Vittorio Ferrari.
\newblock Connecting vision and language with localized narratives.
\newblock In \emph{ECCV}, 2020.

\bibitem[Qi et~al.(2022)Qi, Gao, Hu, Wang, Liu, Bai, Belongie, Yuille, Torr,
  and Bai]{qi2022ijcv_ovis}
Jiyang Qi, Yan Gao, Yao Hu, Xinggang Wang, Xiaoyu Liu, Xiang Bai, Serge
  Belongie, Alan Yuille, Philip H.~S. Torr, and Song Bai.
\newblock Occluded video instance segmentation: A benchmark.
\newblock \emph{IJCV}, 2022.

\bibitem[Radford et~al.(2021)Radford, Kim, Hallacy, Ramesh, Goh, Agarwal,
  Sastry, Askell, Mishkin, Clark, et~al.]{radford2021learning}
Alec Radford, Jong~Wook Kim, Chris Hallacy, Aditya Ramesh, Gabriel Goh,
  Sandhini Agarwal, Girish Sastry, Amanda Askell, Pamela Mishkin, Jack Clark,
  et~al.
\newblock Learning transferable visual models from natural language
  supervision.
\newblock In \emph{ICLR}, 2021.

\bibitem[Raffel et~al.(2020)Raffel, Shazeer, Roberts, Lee, Narang, Matena,
  Zhou, Li, and Liu]{raffel2020exploring}
Colin Raffel, Noam Shazeer, Adam Roberts, Katherine Lee, Sharan Narang, Michael
  Matena, Yanqi Zhou, Wei Li, and Peter~J Liu.
\newblock Exploring the limits of transfer learning with a unified text-to-text
  transformer.
\newblock \emph{JMLR}, 2020.

\bibitem[Ravi et~al.(2024)Ravi, Gabeur, Hu, Hu, Ryali, Ma, Khedr, R{\"a}dle,
  Rolland, Gustafson, et~al.]{ravi2024sam}
Nikhila Ravi, Valentin Gabeur, Yuan-Ting Hu, Ronghang Hu, Chaitanya Ryali,
  Tengyu Ma, Haitham Khedr, Roman R{\"a}dle, Chloe Rolland, Laura Gustafson,
  et~al.
\newblock Sam 2: Segment anything in images and videos.
\newblock \emph{arXiv:2408.00714}, 2024.

\bibitem[Real et~al.(2017)Real, Shlens, Mazzocchi, Pan, and
  Vanhoucke]{real2017youtubeboundingboxes}
Esteban Real, Jonathon Shlens, Stefano Mazzocchi, Xin Pan, and Vincent
  Vanhoucke.
\newblock Youtube-boundingboxes: A large high-precision human-annotated data
  set for object detection in video.
\newblock In \emph{CVPR}, 2017.

\bibitem[Ronneberger et~al.(2015)Ronneberger, Fischer, and
  Brox]{ronneberger2015u}
Olaf Ronneberger, Philipp Fischer, and Thomas Brox.
\newblock U-net: Convolutional networks for biomedical image segmentation.
\newblock In \emph{MICCAI}, 2015.

\bibitem[Russakovsky et~al.(2015)Russakovsky, Deng, Su, Krause, Satheesh, Ma,
  Huang, Karpathy, Khosla, Bernstein, Berg, and
  Fei-Fei]{russakovsky2015imagenet}
Olga Russakovsky, Jia Deng, Hao Su, Jonathan Krause, Sanjeev Satheesh, Sean Ma,
  Zhiheng Huang, Andrej Karpathy, Aditya Khosla, Michael Bernstein,
  Alexander~C. Berg, and Li Fei-Fei.
\newblock Imagenet large scale visual recognition challenge.
\newblock \emph{IJCV}, 2015.

\bibitem[Ryali et~al.(2023)Ryali, Hu, Bolya, Wei, Fan, Huang, Aggarwal,
  Chowdhury, Poursaeed, Hoffman, et~al.]{ryali2023hiera}
Chaitanya Ryali, Yuan-Ting Hu, Daniel Bolya, Chen Wei, Haoqi Fan, Po-Yao Huang,
  Vaibhav Aggarwal, Arkabandhu Chowdhury, Omid Poursaeed, Judy Hoffman, et~al.
\newblock Hiera: A hierarchical vision transformer without the
  bells-and-whistles.
\newblock In \emph{ICLR}, 2023.

\bibitem[Ryoo et~al.(2021)Ryoo, Piergiovanni, Arnab, Dehghani, and
  Angelova]{ryoo2021tokenlearner}
Michael Ryoo, AJ Piergiovanni, Anurag Arnab, Mostafa Dehghani, and Anelia
  Angelova.
\newblock Tokenlearner: Adaptive space-time tokenization for videos.
\newblock \emph{NeurIPS}, 2021.

\bibitem[Seo et~al.(2020)Seo, Lee, and Han]{seo2020urvos}
Seonguk Seo, Joon-Young Lee, and Bohyung Han.
\newblock Urvos: Unified referring video object segmentation network with a
  large-scale benchmark.
\newblock In \emph{ECCV}, 2020.

\bibitem[Shang et~al.(2019)Shang, Di, Xiao, Cao, Yang, and
  Chua]{shang2019icmr_vidor}
Xindi Shang, Donglin Di, Junbin Xiao, Yu Cao, Xun Yang, and Tat-Seng Chua.
\newblock Annotating objects and relations in user-generated videos.
\newblock In \emph{ICMR}, 2019.

\bibitem[Shao et~al.(2022)Shao, Han, Marnerides, and
  Debattista]{shao2022nnls_densecap}
Zhuang Shao, Jungong Han, Demetris Marnerides, and Kurt Debattista.
\newblock Region-object relation-aware dense captioning via transformer.
\newblock \emph{IEEE Transactions on Neural Networks and Learning Systems},
  2022.

\bibitem[Shtedritski et~al.(2023)Shtedritski, Rupprecht, and
  Vedaldi]{shtedritski2023does}
Aleksandar Shtedritski, Christian Rupprecht, and Andrea Vedaldi.
\newblock What does clip know about a red circle? visual prompt engineering for
  vlms.
\newblock In \emph{ICCV}, 2023.

\bibitem[Sun et~al.(2024)Sun, Zhou, Zhao, Yuan, Seybold, Hendon, Schroff, Ross,
  Adam, Hu, et~al.]{sun2024video}
Jennifer~J Sun, Hao Zhou, Long Zhao, Liangzhe Yuan, Bryan Seybold, David
  Hendon, Florian Schroff, David~A Ross, Hartwig Adam, Bo Hu, et~al.
\newblock Video foundation models for animal behavior analysis.
\newblock \emph{bioRxiv}, 2024.

\bibitem[Sun et~al.(2020)Sun, Kretzschmar, Dotiwalla, Chouard, Patnaik, Tsui,
  Guo, Zhou, Chai, Caine, et~al.]{sun2020scalability}
Pei Sun, Henrik Kretzschmar, Xerxes Dotiwalla, Aurelien Chouard, Vijaysai
  Patnaik, Paul Tsui, James Guo, Yin Zhou, Yuning Chai, Benjamin Caine, et~al.
\newblock Scalability in perception for autonomous driving: Waymo open dataset.
\newblock In \emph{CVPR}, 2020.

\bibitem[Team et~al.(2024{\natexlab{a}})Team, Mesnard, Hardin, Dadashi,
  Bhupatiraju, Pathak, Sifre, Rivi{\`e}re, Kale, Love, et~al.]{team2024gemma}
Gemma Team, Thomas Mesnard, Cassidy Hardin, Robert Dadashi, Surya Bhupatiraju,
  Shreya Pathak, Laurent Sifre, Morgane Rivi{\`e}re, Mihir~Sanjay Kale,
  Juliette Love, et~al.
\newblock Gemma: Open models based on gemini research and technology.
\newblock \emph{arXiv:2403.08295}, 2024{\natexlab{a}}.

\bibitem[Team et~al.(2024{\natexlab{b}})Team, Riviere, Pathak, Sessa, Hardin,
  Bhupatiraju, Hussenot, Mesnard, Shahriari, Ram{\'e}, et~al.]{team2024gemma2}
Gemma Team, Morgane Riviere, Shreya Pathak, Pier~Giuseppe Sessa, Cassidy
  Hardin, Surya Bhupatiraju, L{\'e}onard Hussenot, Thomas Mesnard, Bobak
  Shahriari, Alexandre Ram{\'e}, et~al.
\newblock Gemma 2: Improving open language models at a practical size.
\newblock \emph{arXiv:2408.00118}, 2024{\natexlab{b}}.

\bibitem[Touvron et~al.(2023)Touvron, Lavril, Izacard, Martinet, Lachaux,
  Lacroix, Rozi{\`e}re, Goyal, Hambro, Azhar, et~al.]{touvron2023llama}
Hugo Touvron, Thibaut Lavril, Gautier Izacard, Xavier Martinet, Marie-Anne
  Lachaux, Timoth{\'e}e Lacroix, Baptiste Rozi{\`e}re, Naman Goyal, Eric
  Hambro, Faisal Azhar, et~al.
\newblock Llama: Open and efficient foundation language models.
\newblock \emph{arXiv:2302.13971}, 2023.

\bibitem[Vaswani et~al.(2017)Vaswani, Shazeer, Parmar, Uszkoreit, Jones, Gomez,
  Kaiser, and Polosukhin]{vaswani2017attention}
Ashish Vaswani, Noam Shazeer, Niki Parmar, Jakob Uszkoreit, Llion Jones,
  Aidan~N Gomez, {\L}ukasz Kaiser, and Illia Polosukhin.
\newblock Attention is all you need.
\newblock In \emph{NeurIPS}, 2017.

\bibitem[Vedantam et~al.(2015)Vedantam, Lawrence~Zitnick, and
  Parikh]{vedantam2015cider}
Ramakrishna Vedantam, C Lawrence~Zitnick, and Devi Parikh.
\newblock Cider: Consensus-based image description evaluation.
\newblock In \emph{CVPR}, 2015.

\bibitem[Villegas et~al.(2022)Villegas, Babaeizadeh, Kindermans, Moraldo,
  Zhang, Saffar, Castro, Kunze, and Erhan]{villegas2022phenaki}
Ruben Villegas, Mohammad Babaeizadeh, Pieter-Jan Kindermans, Hernan Moraldo,
  Han Zhang, Mohammad~Taghi Saffar, Santiago Castro, Julius Kunze, and Dumitru
  Erhan.
\newblock Phenaki: Variable length video generation from open domain textual
  descriptions.
\newblock In \emph{ICLR}, 2022.

\bibitem[Voigtlaender et~al.(2023)Voigtlaender, Changpinyo, Pont-Tuset,
  Soricut, and Ferrari]{voigtlaender2023connecting}
Paul Voigtlaender, Soravit Changpinyo, Jordi Pont-Tuset, Radu Soricut, and
  Vittorio Ferrari.
\newblock Connecting vision and language with video localized narratives.
\newblock In \emph{CVPR}, 2023.

\bibitem[Wang et~al.(2022)Wang, Yang, Hu, Li, Lin, Gan, Liu, Liu, and
  Wang]{wang2022git}
Jianfeng Wang, Zhengyuan Yang, Xiaowei Hu, Linjie Li, Kevin Lin, Zhe Gan,
  Zicheng Liu, Ce Liu, and Lijuan Wang.
\newblock Git: A generative image-to-text transformer for vision and language.
\newblock \emph{TMLR}, 2022.

\bibitem[Wang et~al.(2021{\natexlab{a}})Wang, Zhang, Lu, Zheng, Cheng, and
  Luo]{wang2021iccv_densevidcap}
Teng Wang, Ruimao Zhang, Zhichao Lu, Feng Zheng, Ran Cheng, and Ping Luo.
\newblock End-to-end dense video captioning with parallel decoding.
\newblock In \emph{ICCV}, 2021{\natexlab{a}}.

\bibitem[Wang et~al.(2023{\natexlab{a}})Wang, Zhang, Fei, Ge, Zheng, Tang, Li,
  Gao, Zhao, Shan, and Zheng]{wang2023caption}
Teng Wang, Jinrui Zhang, Junjie Fei, Yixiao Ge, Hao Zheng, Yunlong Tang, Zhe
  Li, Mingqi Gao, Shanshan Zhao, Ying Shan, and Feng Zheng.
\newblock Caption anything: Interactive image description with diverse
  multimodal controls.
\newblock \emph{arXiv:2305.02677}, 2023{\natexlab{a}}.

\bibitem[Wang et~al.(2021{\natexlab{b}})Wang, Feiszli, Wang, and
  Tran]{wang2021uvo}
Weiyao Wang, Matt Feiszli, Heng Wang, and Du Tran.
\newblock Unidentified video objects: A benchmark for dense, open-world
  segmentation.
\newblock In \emph{CVPR}, 2021{\natexlab{b}}.

\bibitem[Wang et~al.(2023{\natexlab{b}})Wang, Shi, Li, Wang, Huang, Xing, Chen,
  Li, Zhu, Cao, et~al.]{wang2023all}
Weiyun Wang, Min Shi, Qingyun Li, Wenhai Wang, Zhenhang Huang, Linjie Xing, Zhe
  Chen, Hao Li, Xizhou Zhu, Zhiguo Cao, et~al.
\newblock The all-seeing project: Towards panoptic visual recognition and
  understanding of the open world.
\newblock \emph{ICLR}, 2023{\natexlab{b}}.

\bibitem[Wang et~al.(2024{\natexlab{a}})Wang, Ren, Luo, Li, Yan, Chen, Wang,
  Li, Lu, Zhu, et~al.]{wang2024all}
Weiyun Wang, Yiming Ren, Haowen Luo, Tiantong Li, Chenxiang Yan, Zhe Chen,
  Wenhai Wang, Qingyun Li, Lewei Lu, Xizhou Zhu, et~al.
\newblock The all-seeing project v2: Towards general relation comprehension of
  the open world.
\newblock \emph{arXiv:2402.19474}, 2024{\natexlab{a}}.

\bibitem[Wang et~al.(2024{\natexlab{b}})Wang, Darrell, Rambhatla, Girdhar, and
  Misra]{wang2024instancediffusion}
Xudong Wang, Trevor Darrell, Sai~Saketh Rambhatla, Rohit Girdhar, and Ishan
  Misra.
\newblock Instancediffusion: Instance-level control for image generation.
\newblock In \emph{CVPR}, 2024{\natexlab{b}}.

\bibitem[Wu et~al.(2022)Wu, Jiang, Sun, Yuan, and Luo]{wu2022language}
Jiannan Wu, Yi Jiang, Peize Sun, Zehuan Yuan, and Ping Luo.
\newblock Language as queries for referring video object segmentation.
\newblock In \emph{CVPR}, 2022.

\bibitem[Wu et~al.(2023)Wu, Jiang, Yan, Lu, Yuan, and Luo]{wu2023uniref++}
Jiannan Wu, Yi Jiang, Bin Yan, Huchuan Lu, Zehuan Yuan, and Ping Luo.
\newblock Uniref++: Segment every reference object in spatial and temporal
  spaces.
\newblock \emph{arXiv:2312.15715}, 2023.

\bibitem[Wu et~al.(2024{\natexlab{a}})Wu, Jiang, Liu, Yuan, Bai, and
  Bai]{wu2024general}
Junfeng Wu, Yi Jiang, Qihao Liu, Zehuan Yuan, Xiang Bai, and Song Bai.
\newblock General object foundation model for images and videos at scale.
\newblock In \emph{CVPR}, 2024{\natexlab{a}}.

\bibitem[Wu et~al.(2024{\natexlab{b}})Wu, Wang, Yang, Gan, Liu, Yuan, and
  Wang]{wu2024grit}
Jialian Wu, Jianfeng Wang, Zhengyuan Yang, Zhe Gan, Zicheng Liu, Junsong Yuan,
  and Lijuan Wang.
\newblock Grit: A generative region-to-text transformer for object
  understanding.
\newblock \emph{ECCV}, 2024{\natexlab{b}}.

\bibitem[Wu et~al.(2024{\natexlab{c}})Wu, Wang, Tang, Wu, He, Ouyang, Torr, and
  Wu]{wu2024eccv_dettoolchain}
Yixuan Wu, Yizhou Wang, Shixiang Tang, Wenhao Wu, Tong He, Wanli Ouyang, Philip
  Torr, and Jian Wu.
\newblock Dettoolchain: A new prompting paradigm to unleash detection ability
  of mllm.
\newblock \emph{ECCV}, 2024{\natexlab{c}}.

\bibitem[Xu et~al.(2024)Xu, Zhou, Yan, Gu, Arnab, Sun, Wang, and
  Schmid]{xu2024cvpr_pixelllm}
Jiarui Xu, Xingyi Zhou, Shen Yan, Xiuye Gu, Anurag Arnab, Chen Sun, Xiaolong
  Wang, and Cordelia Schmid.
\newblock Pixel-aligned language model.
\newblock In \emph{CVPR}, 2024.

\bibitem[Xu et~al.(2018)Xu, Yang, Fan, Yang, Yue, Liang, Price, Cohen, and
  Huang]{xu2018youtube}
Ning Xu, Linjie Yang, Yuchen Fan, Jianchao Yang, Dingcheng Yue, Yuchen Liang,
  Brian Price, Scott Cohen, and Thomas Huang.
\newblock Youtube-vos: Sequence-to-sequence video object segmentation.
\newblock In \emph{ECCV}, 2018.

\bibitem[Xue et~al.(2024)Xue, Shu, Awadalla, Wang, Yan, Purushwalkam, Zhou,
  Prabhu, Dai, Ryoo, et~al.]{xue2024xgen}
Le Xue, Manli Shu, Anas Awadalla, Jun Wang, An Yan, Senthil Purushwalkam,
  Honglu Zhou, Viraj Prabhu, Yutong Dai, Michael~S Ryoo, et~al.
\newblock xgen-mm (blip-3): A family of open large multimodal models.
\newblock \emph{arXiv preprint arXiv:2408.08872}, 2024.

\bibitem[Yang et~al.(2023{\natexlab{a}})Yang, Nagrani, Seo, Miech, Pont-Tuset,
  Laptev, Sivic, and Schmid]{yang2023cvpr_vid2seq}
Antoine Yang, Arsha Nagrani, Paul~Hongsuck Seo, Antoine Miech, Jordi
  Pont-Tuset, Ivan Laptev, Josef Sivic, and Cordelia Schmid.
\newblock Vid2seq: Large-scale pretraining of a visual language model for dense
  video captioning.
\newblock In \emph{CVPR}, 2023{\natexlab{a}}.

\bibitem[Yang et~al.(2023{\natexlab{b}})Yang, Gao, Li, Gao, Wang, and
  Zheng]{yang2023track}
Jinyu Yang, Mingqi Gao, Zhe Li, Shang Gao, Fangjing Wang, and Feng Zheng.
\newblock Track anything: Segment anything meets videos.
\newblock \emph{arXiv:2304.11968}, 2023{\natexlab{b}}.

\bibitem[Yang et~al.(2023{\natexlab{c}})Yang, Zhang, Li, Zou, Li, and
  Gao]{yang2023arxiv_setofmark}
Jianwei Yang, Hao Zhang, Feng Li, Xueyan Zou, Chunyuan Li, and Jianfeng Gao.
\newblock Set-of-mark prompting unleashes extraordinary visual grounding in
  gpt-4v.
\newblock \emph{arXiv preprint arXiv:2310.11441}, 2023{\natexlab{c}}.

\bibitem[Yang and Yang(2022)]{yang2022deaot}
Zongxin Yang and Yi Yang.
\newblock Decoupling features in hierarchical propagation for video object
  segmentation.
\newblock In \emph{NeurIPS}, 2022.

\bibitem[Yang et~al.(2021)Yang, Wei, and Yang]{yang2021aot}
Zongxin Yang, Yunchao Wei, and Yi Yang.
\newblock Associating objects with transformers for video object segmentation.
\newblock In \emph{NeurIPS}, 2021.

\bibitem[Yang et~al.(2024)Yang, Miao, Wei, Wang, Wang, and Yang]{yang2021aost}
Zongxin Yang, Jiaxu Miao, Yunchao Wei, Wenguan Wang, Xiaohan Wang, and Yi Yang.
\newblock Scalable video object segmentation with identification mechanism.
\newblock \emph{TPAMI}, 2024.

\bibitem[Yao et~al.(2015)Yao, Torabi, Cho, Ballas, Pal, Larochelle, and
  Courville]{yao2015iccv_videocaption}
Li Yao, Atousa Torabi, Kyunghyun Cho, Nicolas Ballas, Christopher Pal, Hugo
  Larochelle, and Aaron Courville.
\newblock Describing videos by exploiting temporal structure.
\newblock In \emph{ICCV}, 2015.

\bibitem[Yu et~al.(2016)Yu, Poirson, Yang, Berg, and Berg]{yu2016modeling}
Licheng Yu, Patrick Poirson, Shan Yang, Alexander~C Berg, and Tamara~L Berg.
\newblock Modeling context in referring expressions.
\newblock In \emph{ECCV}, 2016.

\bibitem[Yuan et~al.(2024)Yuan, Li, Liu, Tang, Luo, Qin, Zhang, and
  Zhu]{yuan2024osprey}
Yuqian Yuan, Wentong Li, Jian Liu, Dongqi Tang, Xinjie Luo, Chi Qin, Lei Zhang,
  and Jianke Zhu.
\newblock Osprey: Pixel understanding with visual instruction tuning.
\newblock In \emph{CVPR}, 2024.

\bibitem[Zhang et~al.(2023)Zhang, Sun, Chen, Xiao, Shao, Zhang, Liu, Chen, and
  Luo]{zhang2023gpt4roi}
Shilong Zhang, Peize Sun, Shoufa Chen, Min Xiao, Wenqi Shao, Wenwei Zhang, Yu
  Liu, Kai Chen, and Ping Luo.
\newblock Gpt4roi: Instruction tuning large language model on
  region-of-interest.
\newblock \emph{arXiv:2307.03601}, 2023.

\bibitem[Zhang et~al.(2020)Zhang, Zhao, Zhao, Wang, Liu, and
  Gao]{zhang2020does}
Zhu Zhang, Zhou Zhao, Yang Zhao, Qi Wang, Huasheng Liu, and Lianli Gao.
\newblock Where does it exist: Spatio-temporal video grounding for multi-form
  sentences.
\newblock In \emph{CVPR}, 2020.

\bibitem[Zheng et~al.(2024)Zheng, Gou, Kil, Sun, and
  Su]{zheng2024icml_grounded_gpt}
Boyuan Zheng, Boyu Gou, Jihyung Kil, Huan Sun, and Yu Su.
\newblock Gpt-4v (ision) is a generalist web agent, if grounded.
\newblock \emph{ICML}, 2024.

\bibitem[Zheng et~al.(2021)Zheng, Lu, Zhao, Zhu, Luo, Wang, Fu, Feng, Xiang,
  Torr, et~al.]{zheng2021rethinking}
Sixiao Zheng, Jiachen Lu, Hengshuang Zhao, Xiatian Zhu, Zekun Luo, Yabiao Wang,
  Yanwei Fu, Jianfeng Feng, Tao Xiang, Philip~HS Torr, et~al.
\newblock Rethinking semantic segmentation from a sequence-to-sequence
  perspective with transformers.
\newblock In \emph{CVPR}, 2021.

\bibitem[Zhou et~al.(2018{\natexlab{a}})Zhou, Louis, and
  Corso]{zhou2018youcook2bb}
Luowei Zhou, Nathan Louis, and Jason~J Corso.
\newblock Weakly-supervised video object grounding from text by loss weighting
  and object interaction.
\newblock In \emph{BMVC}, 2018{\natexlab{a}}.

\bibitem[Zhou et~al.(2018{\natexlab{b}})Zhou, Xu, and Corso]{zhou2018youcook2}
Luowei Zhou, Chenliang Xu, and Jason~J Corso.
\newblock Towards automatic learning of procedures from web instructional
  videos.
\newblock In \emph{AAAI}, 2018{\natexlab{b}}.

\bibitem[Zhou et~al.(2019)Zhou, Kalantidis, Chen, Corso, and
  Rohrbach]{zhou2019grounded}
Luowei Zhou, Yannis Kalantidis, Xinlei Chen, Jason~J Corso, and Marcus
  Rohrbach.
\newblock Grounded video description.
\newblock In \emph{CVPR}, 2019.

\bibitem[Zhou et~al.(2023)Zhou, Arnab, Sun, and Schmid]{zhou2023dense}
Xingyi Zhou, Anurag Arnab, Chen Sun, and Cordelia Schmid.
\newblock Dense video object captioning from disjoint supervision.
\newblock \emph{arXiv:2306.11729}, 2023.

\bibitem[Zulfikar et~al.(2024)Zulfikar, Mahadevan, Voigtlaender, and
  Leibe]{zulfikar2024cvpr_pointvos}
Idil~Esen Zulfikar, Sabarinath Mahadevan, Paul Voigtlaender, and Bastian Leibe.
\newblock Point-{VOS}: Pointing up video object segmentation.
\newblock In \emph{CVPR}, 2024.

\end{thebibliography}
